\pdfoutput=1 % tell arXiv to build with pdfLaTeX
\documentclass{article}

    \PassOptionsToPackage{numbers, compress}{natbib}
\usepackage[main, final]{neurips_2026}

\usepackage[utf8]{inputenc} % allow utf-8 input
\usepackage[T1]{fontenc}    % use 8-bit T1 fonts
\usepackage{hyperref}       % hyperlinks
\usepackage{url}            % simple URL typesetting
\usepackage{booktabs}       % professional-quality tables
\usepackage{array}
\usepackage{amsfonts}       % blackboard math symbols
\usepackage{nicefrac}       % compact symbols for 1/2, etc.
\usepackage{microtype}      % microtypography
\usepackage{xcolor}         % colors

\usepackage{xspace} % auto spacing after newcommand
\usepackage{amsmath}
\usepackage{float} % To force placing tables "H"ere
\usepackage{multirow}
\usepackage[normalem]{ulem} % For strikethrough; Use [normalem] to prevent underlining emphasis

\usepackage[disable]{todonotes}

\usepackage{titlesec}
\titlespacing*{\paragraph}{0pt}{0ex}{1em} % 0ex: No extra spacing before \paragraph
\titlespacing*{\section}
  {0pt}  % Left indentation
  {6pt} % Space before the title (above)
  {5pt}  % Space after the title (below)
\titlespacing*{\subsection}
  {0pt}  % Left indentation
  {6pt} % Space before the title (above)
  {5pt}  % Space after the title (below)
\renewcommand{\arraystretch}{0.85}

\usepackage{afterpage}
\newcommand{\ours}{TraXion}
\newcommand{\sig}{{\tiny{*}}}
\usepackage{graphicx}
\newcommand{\std}[1]{\scalebox{0.8}{$\scriptscriptstyle\pm#1$}}
\newcommand{\hc}[1]{\multicolumn{1}{c}{#1}}  % centered header over a left-aligned column

\title{\ours: Rethinking Pre-training Frameworks for Mobility and Beyond}

\author{%
  Shang-Ling Hsu \\
  University of Southern California \\
  Los Angeles, California, USA \\
  \texttt{hsushang@usc.edu} \\
  \And
  Mark Tenzer \\
  Novateur Research Solutions \\
  Ashburn, Virginia, USA \\
  \texttt{mtenzer@novateur.ai} \\
  \AND
  Cyrus Shahabi \\
  University of Southern California \\
  Los Angeles, California, USA \\
  \texttt{shahabi@usc.edu} \\
  \And
  Khurram Shafique \\
  Novateur Research Solutions \\
  Ashburn, Virginia, USA \\
  \texttt{kshafique@novateur.ai} \\
}

\begin{document}

\maketitle

\begin{abstract}

Human mobility differs from text and from generic time series in three structural ways: visits are tuple-valued events whose meaning depends on the joint distribution over location, time, and activity; users carry persistent signatures across trajectories; and visits are not independent across users, since co-location at shared places is a primary signal. Existing pre-training recipes for mobility import objectives from language modeling, treating trajectories as sentences and visits as tokens, an analogy that fails against each of the three properties above. These properties define a broader class, multi-entity spatiotemporal event streams (MESES), spanning enterprise authentication logs, electronic health records, and other event-stream domains where entities share infrastructure, schedules, or contexts. We make the properties precise as three axioms that any pre-training framework for MESES should satisfy, and introduce TraXion, whose objectives and architecture are jointly designed to meet them. A single TraXion checkpoint per dataset leads task-specific baselines on every location-ranking and detection metric across six public mobility datasets covering anomaly detection, next-POI recommendation, next-visit prediction, and social-link prediction. The same recipe, applied unchanged to enterprise authentication logs and ICU mortality prediction, matches or exceeds prior work on both, showing that event streams from domains as different as mobility, security, and healthcare can be modeled under a single framework.\footnote{Code: \url{https://github.com/ktxlh/TraXion}}

\end{abstract}

\section{Introduction}
\label{sec:intro}
Across mobility, security, healthcare, and many other domains, a common kind of data appears: multiple entities generate sequences of discrete events, where each event is a tuple over space, time, and activity, and where the entities are coupled through the contexts in which their events occur. Human mobility is the canonical example. Users visit points of interest, dwell for a time, and meet at the same places as family, coworkers, and friends. The same structure appears in many event-stream domains where multiple entities share infrastructure, schedules, or contexts. For example, in enterprise authentication, users access shared hosts under role-conditioned schedules; in electronic health records, ICU stays draw from a shared catalog of clinical actions (medications, lab tests, treatments). We call this class of data {\it multi-entity spatiotemporal event streams} ({\bf MESES}). 

Self-supervised pre-training is the natural way to build general-purpose representations from MESES data, but the dominant pre-training frameworks were designed for language and fail on MESES in specific, characterizable
ways. Autoregressive and masked token prediction, the recipes that drive most modern sequence models for MESES, were developed under a linguistic analogy: events are like words, and entity-trajectories are like sentences ~\citep{siampou2026mepois, zhu2025unitraj, gong2024mobilityllm, li2026trajflow, guo2021logbert, han2023loggpt, renc2024ethos}. Three properties of MESES data, however, distinguish events from language tokens at a structural level. First, a token's identity is fully captured by a vocabulary index, but a MESES event is a tuple over location, time, duration, and activity, and its meaning depends on the joint distribution over these axes: the same activity at different times, or at the same context by different entities, is not the same event. Second, a sentence is drawn from a shared linguistic process, but each entity in MESES carries a persistent signature that a token-level model has no direct mechanism to represent. Third, sentences are largely independent across speakers, but MESES events are explicitly not independent across entities, because co-occurrence at shared contexts is a primary signal rather than noise. A pre-training framework that ignores these three properties cannot model MESES data faithfully.

This paper makes three contributions. First, we define MESES as a class of data unified by tuple-valued events, persistent entity signatures, and shared-context structure, and formalize these as three axioms that any pre-training framework for MESES should satisfy (\S\ref{sec:axioms}). Second, we introduce \textbf{\ours}, a pretraining framework that couples a noise-detection objective and an entity-prototype contrastive objective on a factorized Transformer backbone with sequence, feature, and co-occurrence attention axes, satisfying the three axioms (\S\ref{sec:method}). Third, we evaluate \ours\ on three MESES instances spanning mobility, enterprise security, and clinical event data (\S\ref{sec:experiments}). On six public mobility datasets covering anomaly detection, next-POI recommendation, next-visit prediction, and social-link prediction, a single \ours\ checkpoint per dataset leads specialized task-specific baselines on every location-ranking and detection metric, with visit-level anomaly AP improving by $12$ points over the strongest baseline. On the two non-mobility MESES instances, \ours\ transfers without architectural change: scored without fine-tuning on the LANL enterprise authentication log, it attains the highest mean on four of six metrics and is statistically indistinguishable from five dedicated log-anomaly detectors, and on eICU-CRD and MIMIC-IV it leads seven of eight mortality-prediction columns. 
Together, these contributions identify a pre-training recipe shaped by what MESES data demands rather than what language modeling assumes, transferring across mobility, security, and healthcare.
\section{Related work}
\label{sec:related-work}
Three nearby data classes might appear to subsume MESES. It is not a
generic multivariate time series: events are sparse and discrete in
time rather than sampled at regular intervals. It is not a generic
event stream: events are coupled across entities through shared
contexts. And it is not a sequence-of-tokens domain: events are
tuple-valued rather than atomic. We position \ours\ against each in
turn.
\paragraph{Marked temporal point processes and irregular multivariate time series.}
Standard MTPP families assume neither a shared metric context set nor population-scale entity coupling. Multi-dimensional Hawkes processes~\citep{zhou2013learning, linderman2014discovering} and Dirichlet-Hawkes mixtures~\citep{du2015dirichlet} couple streams pairwise or through clustering priors and scale super-linearly in the number of entities. Neural-MTPP variants~\citep{du2016rmptt, mei2017nhp, zuo2020thp, yang2022anhp, shchur2021review} and irregular-MTS models (GRU-D~\citep{che2018grud}, mTAN~\citep{shukla2021mtan}, TFT~\citep{lim2021tft}) target intensity estimation or forecasting under proper likelihoods.  \ours{} pre-trains representations for downstream classification and ranking and is not a substitute on likelihood tasks (next-event time, hazard, return-time).

\paragraph{Self-supervised pre-training for sequences.}
Our noise detection objective ports ELECTRA's replaced-token-detection~\citep{clark2020electra} from atomic tokens to tuple-valued events.  Masked-modeling alternatives reconstruct corrupted positions (BERT~\citep{devlin2019bert}, MAE~\citep{he2022mae}, TST~\citep{zerveas2021tst}). TS2Vec~\citep{yue2022ts2vec}, SimMTM~\citep{dong2023simmtm}, and PatchTST~\citep{nie2023patchtst} extend the recipe for time series.  Sequential recommenders (SASRec~\citep{kang2018sasrec}, BERT4Rec~\citep{sun2019bert4rec}, S\textsuperscript{3}-Rec~\citep{zhou2020s3rec}, CL4SRec~\citep{xie2022cl4srec}) share an item embedding table between input and supervision; \ours{} specializes the construction by making the positive class an \emph{entity} rather than an item, the structural commitment that follows from persistent entity signatures.  The closest prior art on this point is trajectory-user-linking methods~\citep{zhou2018trajectory, chen2022mutual, chen2024trajectory, zhang2025scalable}, which supervise on entity identity directly but does not retain a learned per-entity prototype that re-enters the input.  Earlier trajectory-specific pre-training (CTLE~\citep{lin2021ctle}, TALE~\citep{wan2021tale}, TrajCL~\citep{chang2023trajcl}) is closest in input modality.
ME-POIs~\citep{siampou2026mepois} applies the same construction --- a learned prototype that serves as the supervision target --- on the context side, learning prototype representations of POIs from mobility traces.

\paragraph{Foundation models for mobility, urban, and event streams.}
Trajectory foundation models target input modalities other than POI-visit streams: UniTraj~\citep{zhu2025unitraj} pre-trains on billion-scale dense GPS with self-supervised trajectory masking, TrajFlow~\citep{li2026trajflow} flow-matches GPS coordinates, and MobilityGPT~\citep{haydari2025mobilitygpt} autoregressively generates road-link sequences;
we benchmark UniTraj as a representative cross-modality baseline.
UniST~\citep{yuan2024unist} pre-trains across urban tasks on grid-structured spatio-temporal data. AgentMove~\citep{feng2025agentmove}, Mobility-LLM~\citep{gong2024mobilityllm}, and GPT4TS~\citep{zhou2023gpt4ts} prompt or freeze pre-trained text models rather than pre-training from event data, which decouples the representation from the shared-context structure characteristic of MESES. CoBAD~\citep{wen2025cobad} models collective behavior across entities, the closest prior art to \ours's co-occurrence axis, but couples streams through a per-event likelihood at training time rather than a pre-training mechanism, so
the cross-entity signal cannot be reused across downstream tasks. In adjacent domains, BEHRT~\citep{li2020behrt}, Med-BERT~\citep{rasmy2021medbert}, CEHR-BERT~\citep{pang2021cehrbert}, and Ethos~\citep{renc2024ethos} pre-train on EHR event sequences, and LogBERT~\citep{guo2021logbert}, DeepLog~\citep{du2017deeplog}, and Tuor-AAAI17~\citep{tuor2017deep} on enterprise log streams;  all process each entity's stream independently. 

\ours\ differs from prior work along two axes. Supervision decomposes into self-supervised pre-training and a supervised fine-tune, rather than a single end-to-end stage. Cross-entity coupling enters through learned attention and a shared prototype table, rather than a pre-committed graph, a hand-built likelihood, or single-stream processing. A per-baseline comparison appears in Appendix~\ref{appendix:related_work_extended}.

\section{Multi-entity spatiotemporal event streams}
\label{sec:meses}

We formalize the data class introduced in \S\ref{sec:intro} and state
three axioms that any pre-training framework over it should satisfy.

\subsection{The MESES schema}
\label{sec:schema}

A MESES dataset is a finite set of entities $\mathcal{U}$ generating events. We index entity $u \in \mathcal{U}$'s events in chronological order, so the $t$-th event of $u$ is a tuple
$e_{u,t} = (u, x_{u,t}, \tau_{u,t}, a_{u,t}, \delta_{u,t})$ over a shared finite context set $\mathcal{X}$ equipped with an embedding $\phi: \mathcal{X} \to \mathbb{R}^{d_x}$, where $x_{u,t} \in \mathcal{X}$ is the event's context, $\tau_{u,t}$ is its timestamp, $a_{u,t}$ is an activity category, and $\delta_{u,t}$ is an optional duration ($\delta=\emptyset$ for instantaneous events).
The embedding $\phi$ varies by MESES instance: latitude/longitude in mobility, a UMAP of the user--host access matrix in enterprise authentication (Appendix~\ref{appendix:lanl}), and a UMAP of the patient--context co-occurrence matrix in electronic health records (Appendix~\ref{appendix:eicu}).
Two entities \emph{co-occur} when their events fall at the same context $x \in \mathcal{X}$ within a small temporal window. We call $\mathcal{X}$ the \emph{substrate} of the MESES instance, and the distribution over the substrate for each entity $u$ is its \emph{signature}.

\subsection{Three axioms for MESES pre-training}
\label{sec:axioms}

We state three axioms that any pre-training framework for MESES should
satisfy. Each axiom is motivated by a property of MESES introduced in
\S\ref{sec:schema}, and none is enforced by autoregressive next-event
prediction. We use ``framework'' to mean the joint
specification of pre-training objectives and architectural mechanisms;
neither alone is sufficient to satisfy the axioms.

\paragraph{A1 (joint-attribute density).}
\emph{Let $p(e_{u,t} \mid \mathcal{H}_{u,t})$ denote the conditional distribution
over the event tuple given an encoding $\mathcal{H}_{u,t}$ of the entity's
history. A pre-training framework satisfies A1 if its training signal
depends on the joint distribution over the event's attributes
$(x, \tau, a, \delta)$ rather than on marginal or factorized distributions
over them.} 

An event's meaning is determined by the joint configuration of its attributes: the same
activity at different times, or at the same context by different
entities, is not the same event. A signal consistent only with the
marginals does not penalize configurations that are plausible along
each axis individually but implausible in combination.

\paragraph{A2 (entity-coupled representation).}
\emph{Let $r(e_{u,t})$ denote the per-event representation produced by the
framework's encoder, and $g(u)$ a representation associated with entity
$u$. A pre-training framework satisfies A2 if entity identity enters
$r(e_{u,t})$ as an input feature, $g(u)$ is constrained by a supervisory
signal, and the input-side and supervisory-side entity representations
are coupled through a shared mechanism.}

Each entity carries a persistent signature in its distribution over the
substrate, and downstream tasks that depend on entity identity (anomaly
detection relative to an entity's normal behavior, social-link prediction
over entity pairs) require the representation to retain
entity-distinguishing information. A loss that depends only on local prediction does not penalize representations that lose entity-distinguishing information across events.

\paragraph{A3 (shared context).}
\emph{Let $\mathcal{X}$ be the substrate of a MESES instance. A pre-training
framework satisfies A3 if its representations exploit two forms of
recurrence over $\mathcal{X}$: that contexts are shared across entities,
so different entities visiting the same $x \in \mathcal{X}$ provide
mutually informative signal; and that contexts recur within an entity's
history, so the entity's distribution over $\mathcal{X}$ is itself
informative.}

Co-occurrence at shared contexts carries population-level information invisible to
per-entity sequence modeling, and within-entity recurrence makes learning the entity's signature feasible. 

\paragraph{Why language-derived recipes do not enforce any of the three.}
Both autoregressive next-event prediction and masked-token reconstruction supervise at the token level. Each event's attributes are predicted from local context only, with no constraint on their joint structure (A1), no signal targeting the entity producing them (A2), and no mechanism for one entity to inform another (A3). A model trained either way may incidentally pick up substrate structure through shared parameters, but the objective
neither requires nor measures such learning.
\section{Method}
\label{sec:method}
We instantiate the framework from \S\ref{sec:meses} as \ours, a pre-training recipe that satisfies A1--A3 through two coupled objectives and a factorized backbone. 

\subsection{Overview}
\label{sec:method:overview}

For an entity $u \in \mathcal{U}$ with event sequence
$(e_{u,1}, \ldots, e_{u,T_u})$, \ours\ produces (i) a sequence encoder
$f_\theta$ that maps the sequence to a matrix of contextualized event
embeddings $H_u = f_\theta(e_{u,1}, \ldots, e_{u,T_u}) \in
\mathbb{R}^{T_u \times d}$, and (ii) a shared entity embedding table
$P \in \mathbb{R}^{|\mathcal{U}| \times d_f}$ whose row $p_u$ we call the
\emph{prototype} of entity $u$. We write $\mathcal{H}_{u,t}$ for the
$t$-th row of $H_u$, the contextualized representation of event
$e_{u,t}$. The prototype table $P$ plays a dual role: its rows are
input to $f_\theta$ as per-event entity tokens
(\S\ref{sec:method:arch}) and serve as supervisory targets in a
contrastive loss (\S\ref{sec:method:user_loss}). 

The pre-training loss combines the two objectives:
\begin{equation}
\mathcal{L}(\theta, \psi) = \mathcal{L}_{\text{noise}}(\theta, \psi) +
\gamma \, \mathcal{L}_{\text{prototype}}(\theta),
\label{eq:loss}
\end{equation}
where $\psi$ parametrizes the noise detection head and $\gamma > 0$ balances
the two terms. The noise detection objective teaches the encoder what
plausible event sequences look like for entities over the substrate; the
prototype objective teaches it who the sequences belong to. Neither
objective uses task-specific labels, so \ours\ trains on event
logs from any MESES instance.

\subsection{Noise-detection objective}
\label{sec:method:noise}
The noise detection objective trains the encoder to detect events whose attributes are corrupted relative to the entity $u$'s normal pattern. A stochastic operator $\eta$ perturbs the \emph{context} and \emph{time} of some of $u$'s events; the encoder flags which events were perturbed.
\paragraph{Operator and loss.}
Let $E_u = (e_{u,1}, \ldots, e_{u,T_u})$ be a chronologically ordered window of events for entity $u$. Applied to $E_u$, the operator $\eta$ produces a corrupted window $\tilde{E}_u = \eta(E_u)$ and a per-event binary target $y_u \in \{0,1\}^{T_u}$ with $y_{u,t} = 1$ on perturbed events. With probability $p_{\text{norm}}$ the window is left
unchanged ($\tilde{E}_u = E_u$, $y_u = \mathbf{0}$); otherwise a random fraction of events is flagged, and each one independently receives a context perturbation, a time perturbation, or both (Appendix~\ref{appendix:design_intuitions}). A linear head $g_\psi$ on the contextualized embedding $\mathcal{H}_{u,t} = f_\theta(\tilde{E}_u)_t$ predicts per-event logits trained with masked binary cross-entropy:

\begin{equation}
\mathcal{L}_{\text{noise}} = \frac{1}{|\mathcal{V}|}
\sum_{(u,t) \in \mathcal{V}}
\text{BCE}\!\left(g_\psi(\mathcal{H}_{u,t}),\, y_{u,t}\right),
\label{eq:denoise}
\end{equation}

over the non-padded events $\mathcal{V}$.

\paragraph{Context perturbation.}
An event's context $x_{u,t}$ is resampled by drawing a point uniformly from the bounding box of $\phi(\mathcal{X})$, snapping to the nearest context $x' \in \mathcal{X}$ (the nearest POI in mobility, the nearest host in enterprise authentication), and replacing the event's activity $a_{u,t}$ with the activity associated with $x'$. The perturbed event remains locally realistic in geometry but is disconnected from the entity's typical distribution over the substrate.

\paragraph{Time perturbation.}
When event $B$ is flagged, we resample its timestamp uniformly within the window between its neighbors $A$ and $C$, leaving $B$'s context unchanged.  The new $\tau'_B$ disrupts $B$'s timing without changing the context where the event occurred.

\paragraph{How this addresses A1.}
Detecting whether an event's context or time has been perturbed
requires modeling the joint distribution over the event's attributes: a perturbed context at a typical hour, or a
perturbed time at the entity's typical context, is implausible only
jointly. This choice
makes the objective MESES-universal: context and time are the two
axes every instance shares, so the same operator runs on
mobility, authentication, and clinical events.

\subsection{Entity-prototype contrastive objective}
\label{sec:method:user_loss}

The noise-detection objective trains the encoder to model events; the prototype
objective trains it to remember whose events they are. The supervisory
signal is the entity identifier, which is present in every event of a
MESES dataset and requires no additional labeling.

\paragraph{Operator and loss.}
For each minibatch $B$ of entities, let $\mathcal{V}^+ \subseteq
\mathcal{V}$ denote the unperturbed events ($y_{u,t} = 0$). For each such
event we take the $\ell_2$-normalized event representation
$\bar{\mathcal{H}}_{u,t}$ and the $\ell_2$-normalized prototype
$\bar{p}_u$, and minimize the InfoNCE loss \citep{oord2018cpc}
\begin{equation}
\mathcal{L}_{\text{prototype}} = \frac{1}{|\mathcal{V}^+|}
\sum_{(u,t) \in \mathcal{V}^+}
- \log \frac{\exp\!\left(\bar{\mathcal{H}}_{u,t}^\top \bar{p}_u / \beta\right)}
{\sum_{u' \in B} \exp\!\left(\bar{\mathcal{H}}_{u,t}^\top \bar{p}_{u'} / \beta\right)},
\label{eq:prototype}
\end{equation}
where $\beta$ is a temperature. The numerator pulls each event
representation toward its own entity's prototype; the denominator pushes
it away from the prototypes of the others.

Anchors in Eq.~\ref{eq:prototype} are restricted to unperturbed events ($\mathcal{V}^+$) so the prototype summarizes the entity's normal behavior rather than corrupted events (Appendix~\ref{appendix:anchoring}).

\paragraph{How this addresses A2.}
The prototype table $P$ meets A2's three requirements through a single mechanism: the same vector $p_u$ enters each event representation as an input feature and serves as the supervisory target in Eq.~\ref{eq:prototype}. The contrastive loss therefore shapes $p_u$ to carry exactly the entity-distinguishing information the downstream tasks of \S\ref{sec:axioms} require.

\subsection{Backbone architecture}
\label{sec:method:arch}

\begin{figure}[t]
\centering
\includegraphics[width=\textwidth]{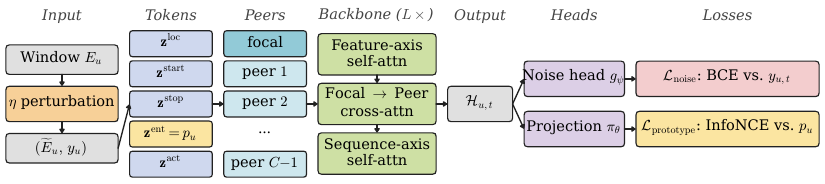}
\caption{Pre-training architecture of \ours{}. Yellow marks the entity prototype $p_u$, which is both an input token and the target of $\mathcal{L}_{\text{prototype}}$. Equations are in Appendix~\ref{appendix:formal}.}
\label{fig:traxion_arch}
\end{figure}

The backbone (Figure~\ref{fig:traxion_arch}) produces the contextualized event representation $\mathcal{H}_{u,t}$ used by both pre-training objectives, and it implements the architectural mechanism through which representations exploit cross-entity context. The first role is shared with most
sequence encoders. The second is what makes the backbone MESES-specific. \ours\ adapts factorized attention over the sequence and feature axes from prior work on multivariate sequence modeling and extends it with a third axis for co-occurrence.

\paragraph{Per-event feature tokens.}\label{sec:method:feature}\label{sec:method:agent}
Each event is represented as $F$ tokens of dimension $d_f = d/F$: a multi-scale Space2Vec~\citep{mai2020space2vec} encoding of $\phi(x_{u,t})$ over $N_s$ scales in $[\lambda_{\min}, \lambda_{\max}]$, two Time2Vec~\citep{kazemi2019time2vec} encodings of $\tau_{u,t}$ and $\tau_{u,t}+\delta_{u,t}$ (collapsed to one when $\delta=0$), the entity prototype $p_u$ looked up from $P$ (factored as $\mathcal{U}\!\to\!h\text{-latent}\!\to\!d_f$ for large instances, $h \ll d_f$; Appendix~\ref{appendix:hyperparameters}), and a linear projection of the activity $a_{u,t}$.

\paragraph{Factorized attention with a co-occurrence axis.}\label{sec:method:attn}
Attention is applied independently along the sequence, feature, and co-occurrence axes of a rank-5 tensor $(B,T,F,C,d_f)$, where $C$ is the size of the co-occurrence window (the focal event plus up to $C{-}1$ peers); factorizing the first two axes is standard for multivariate sequences~\citep{liu2024itransformer,zhang2023crossformer} and reduces the $O((TFC)^2)$ cost of full attention to a sum of standard sub-layers.
The third axis is what makes the backbone MESES-specific: for each focal event we retrieve up to $C{-}1$ temporally proximate events from other entities at the same context $x$ (Appendix~\ref{appendix:co-location}), and the co-occurrence sub-layer attends from the focal event to its peers but not vice versa,
since each peer is itself the focal slot in its own window (Appendix~\ref{appendix:formal:attn}).

\paragraph{How this addresses A3.}
A3 requires exploiting two forms of substrate recurrence: contexts shared across entities, and contexts recurring within an entity's history. The co-occurrence axis addresses the first by making peer events at shared contexts directly available to the focal event. The prototype table $P$, trained over the full corpus by
Eq.~\ref{eq:prototype}, addresses the second by encoding each
entity's distribution over substrate contexts in a fixed vector. The prototype
captures the entity's typical pattern across $\mathcal{X}$, and
the co-occurrence axis captures the population-level pattern at a
specific $x$.

\section{Experiments}
\label{sec:experiments}

\emph{Datasets.}
We evaluate on three MESES instances. For mobility, we use six public human-mobility datasets: NUMOSIM-LA~\citep{stanford2024numosim}, Urban Anomalies-Berlin/-Atlanta (UA)~\citep{amiri2024urban}, Foursquare-Tokyo~\citep{yang2019revisiting}, Gowalla-Stockholm and Gowalla-Austin~\citep{cho2011friendship}. For enterprise authentication, we use a subsampled cohort of the LANL unified host-and-network authentication log~\citep{kent2015comprehensive} ($190{,}944$ events from $1{,}598$ users; Appendix~\ref{appendix:lanl}). For electronic health records, we use PhysioNet eICU-CRD~\citep{pollard2018eicu} ($197{,}307$ ICU stays from $138{,}176$ patients) and MIMIC-IV~\citep{johnson2023mimic} ($94{,}407$ stays from $65{,}360$ patients), both with in-hospital mortality as the label. Per-dataset statistics are in Appendix~\ref{appendix:datasets}. Across these instances we cover four downstream tasks: visit/agent-level anomaly detection (\S\ref{sec:experiments:humob-anomaly}, \S\ref{sec:experiments:auth}), next-POI and next-visit prediction (\S\ref{sec:experiments:next_visit}), social-link inference (\S\ref{sec:experiments:social}), and ICU mortality prediction (\S\ref{sec:experiments:eicu}).

\emph{Protocol.}
For every dataset we pre-train one shared backbone (\S\ref{sec:method:arch}) with the objectives of \S\ref{sec:method:noise}--\ref{sec:method:user_loss}, then fine-tune once per downstream task with a per-task head described in the corresponding subsection.
Whether to bypass the pre-trained co-occurrence sub-layer at fine-tune is decided per seed on the validation criterion: both routes are trained for every seed, and the selected one is reported. Rationale in Appendix~\ref{appendix:dispatch}.

\emph{Metrics.}
For binary classification (anomaly detection, ICU mortality, social-link) we report AP (average precision) and Max F1, with AUROC where applicable. We additionally report Sens@Sp.9 (sensitivity at $90\%$ specificity) for ICU mortality. For ranking (next-POI, next-visit, social-link) we report H@$k$ (hit-rate at $k$) and MRR (mean reciprocal rank).
For next-visit time prediction we report T$\pm60$: for models that emit a density, the predicted probability mass within $60$\,min of the ground-truth check-in time; for point-prediction baselines, the fraction of predictions falling within that window. Both are conditioned on the true next location, following TrajGPT~\citep{hsu2024trajgpt}.

\emph{Implementation and reporting.}
Splits, cohorts, hyperparameters, and the iterative HP search applied identically to \ours{} and retuned baselines are in Appendices~\ref{appendix:implementation}--\ref{appendix:baseline_implementation},~\ref{appendix:hp_search}. Runs use one NVIDIA RTX 6000 Ada (48\,GB); every end-to-end pre-train plus fine-tune completes within $5$ days, with measured per-stage times in Appendix~\ref{appendix:wallclock}. Baselines use released code except UniTraj$^{\dagger}$~\citep{zhu2025unitraj} (adapted; Appendix~\ref{appendix:unitraj_adaptation}), COAST~\citep{xin2025coast} (re-implemented; Appendix~\ref{appendix:poi_baselines}), and some LANL log baselines (re-implemented; Appendix~\ref{appendix:lanl_baselines}).
Tables~\ref{tab:experiments:humob-anomaly}--\ref{tab:experiments:eicu} report mean $\pm$ sample standard deviation; higher is better.
\ours{} is at $5$ seeds in every cell, as is every baseline that leads a column; the remaining baselines are at $3$--$5$ seeds, and EBM is deterministic by construction (single run).
\textbf{Bold} marks the best mean in each column and \underline{underline} the second best.
\sig\ marks $p\!\le\!.05$ wins \emph{or} losses for \ours{} against the strongest per-column baseline (one-sided Wilcoxon rank-sum on the per-seed scores).
Appendix~\ref{appendix:bootstrap} reports percentile bootstrap intervals for the anomaly-detection results of Tables~\ref{tab:experiments:humob-anomaly} and~\ref{tab:experiments:auth}.

\subsection{Human mobility anomaly detection}\label{sec:experiments:humob-anomaly}

Anomaly detection on three simulated benchmarks (ground-truth per-visit/per-agent labels for inserted-visit attacks and behavioral shifts) directly tests the noise-detection objective (\S\ref{sec:method:noise}); UA-Berlin and UA-Atlanta (\texttt{centralized} mode) provide only agent-level labels.
We fine-tune \ours{} with dataset-specific self-supervised objectives matched to each benchmark's anomaly type (Appendix~\ref{appendix:hyperparameters}).

\begin{table}[!htbp]
\centering
\scriptsize
\caption{Human mobility anomaly detection results. UA := Urban Anomalies.}
\label{tab:experiments:humob-anomaly}
\setlength{\tabcolsep}{2.25pt}
\begin{tabular}{lllllllll}
\toprule
& \multicolumn{2}{c}{NUMOSIM-LA (visit)} & \multicolumn{2}{c}{NUMOSIM-LA (agent)} & \multicolumn{2}{c}{UA-Berlin} & \multicolumn{2}{c}{UA-Atlanta} \\
\cmidrule(lr){2-3} \cmidrule(lr){4-5} \cmidrule(lr){6-7} \cmidrule(lr){8-9}
Method & \hc{AP} & \hc{Max F1} & \hc{AP} & \hc{Max F1} & \hc{AP} & \hc{Max F1} & \hc{AP} & \hc{Max F1} \\
\midrule
DeepBayesic~\citep{duan2024back} & $.0002\std{.000}$ & $.0009\std{.001}$ & $.0020\std{.000}$ & $.0045\std{.001}$ & $.2177\std{.036}$ & $\underline{.3014}\std{.034}$ & $.1548\std{.008}$ & $.2396\std{.005}$ \\
CoBAD~\citep{wen2025cobad} & $.0067\std{.002}$ & $.0330\std{.010}$ & $.0143\std{.004}$ & $.0628\std{.010}$ & $.1408\std{.002}$ & $.2236\std{.008}$ & $.1625\std{.003}$ & $.2436\std{.008}$ \\
UniTraj$^{\dagger}$~\citep{zhu2025unitraj} & $.0010\std{.000}$ & $.0084\std{.005}$ & $.0052\std{.001}$ & $.0196\std{.001}$ & $\underline{.2728}\std{.018}$ & $.2805\std{.012}$ & $\underline{.3237}\std{.019}$ & $\underline{.3094}\std{.023}$ \\
ICAD~\citep{azarijoo2025icad} & $\underline{.0775}\std{.057}$ & $\underline{.1382}\std{.062}$ & $\underline{.1455}\std{.085}$ & $\underline{.2282}\std{.088}$ & $.1353\std{.011}$ & $.2361\std{.022}$ & $.1644\std{.021}$ & $.2448\std{.009}$ \\
USTAD~\citep{wen2025uncertainty} & $.0079\std{.001}$ & $.0428\std{.003}$ & $.0208\std{.002}$ & $.0769\std{.007}$ & $.1531\std{.004}$ & $.2364\std{.005}$ & $.2328\std{.008}$ & $.2988\std{.006}$ \\
\textbf{\ours{}} & $\mathbf{.1950}\std{.010}$\sig & $\mathbf{.2901}\std{.009}$\sig & $\mathbf{.5764}\std{.015}$\sig & $\mathbf{.6485}\std{.011}$\sig & $\mathbf{.4132}\std{.014}$\sig & $\mathbf{.4261}\std{.024}$\sig & $\mathbf{.4141}\std{.023}$\sig & $\mathbf{.4214}\std{.022}$\sig \\
\bottomrule
\end{tabular}
\end{table}

\emph{Discussion.}
\ours{} tops all $8$ columns, every one of them significantly, with AP gains of $+.117$/$+.431$ over the strongest baseline (ICAD) at NUMOSIM-LA visit-/agent-level and $+.140$/$+.090$ over UniTraj$^{\dagger}$ on UA-Berlin/-Atlanta.
Which co-occurrence route validation selects tracks task structure: it keeps the pathway on NUMOSIM, where inserted-visit anomalies are focal-vs-crowd mismatches at the place, and bypasses it on UA-Atlanta ($4$ of $5$ seeds), where behavioral shifts at preserved place and time leave co-occurrence uninformative; UA-Berlin splits $3$ active to $2$ bypass (Appendix~\ref{appendix:dispatch}).

\subsection{Next POI and next visit prediction}
\label{sec:experiments:next_visit}

We predict the next visit on three LBSN city subsets, decomposed into location ranking (H@$10$/H@$20$) and time prediction (T$\pm60$); Table~\ref{tab:experiments:next_visit} stacks next-POI (location only) above next-visit baselines (joint location and time).

\begin{table}[!htbp]
\centering
\scriptsize
\caption{Next-POI and next-visit prediction results.}
\label{tab:experiments:next_visit}
\setlength{\tabcolsep}{0.75pt}
\begin{tabular}{llllllllll}
\toprule
& \multicolumn{3}{c}{Foursquare-Tokyo} & \multicolumn{3}{c}{Gowalla-Stockholm} & \multicolumn{3}{c}{Gowalla-Austin} \\
\cmidrule(lr){2-4} \cmidrule(lr){5-7} \cmidrule(lr){8-10}
Method & \hc{H@10} & \hc{H@20} & \hc{T$\pm60$} & \hc{H@10} & \hc{H@20} & \hc{T$\pm60$} & \hc{H@10} & \hc{H@20} & \hc{T$\pm60$} \\
\midrule
\multicolumn{10}{l}{\textit{Next POI recommendation}} \\
UniTraj$^{\dagger}$~\citep{zhu2025unitraj} & $.0020\std{.000}$ & $.0038\std{.001}$ & \multirow{6}{*}{N/A} & $.0182\std{.005}$ & $.0233\std{.005}$ & \multirow{6}{*}{N/A} & $.0089\std{.001}$ & $.0169\std{.002}$ & \multirow{6}{*}{N/A} \\
GETNext~\citep{yang2022getnext} & $.1607\std{.010}$ & $.2118\std{.013}$ &  & $.0584\std{.002}$ & $.0780\std{.003}$ &  & $.1463\std{.012}$ & $.1873\std{.013}$ &  \\
COAST~\citep{xin2025coast} & $.3478\std{.001}$ & $.4109\std{.002}$ &  & $.1388\std{.003}$ & $.1627\std{.003}$ &  & $.1296\std{.001}$ & $.1609\std{.000}$ &  \\
MobTCast~\citep{xue2021mobtcast} & $\underline{.3827}\std{.002}$ & $\underline{.4510}\std{.002}$ &  & $.1943\std{.005}$ & $.2245\std{.006}$ &  & $.1459\std{.004}$ & $.1834\std{.005}$ &  \\
LoTNext~\citep{xu2024taming} & $.3318\std{.002}$ & $.3958\std{.002}$ &  & $\underline{.2362}\std{.002}$ & $\underline{.2786}\std{.001}$ &  & $\underline{.2034}\std{.003}$ & $\underline{.2514}\std{.002}$ &  \\
\textbf{\ours{}} & $\mathbf{.4330}\std{.002}$\sig & $\mathbf{.5248}\std{.002}$\sig &  & $\mathbf{.3237}\std{.003}$\sig & $\mathbf{.3946}\std{.003}$\sig &  & $\mathbf{.2755}\std{.005}$\sig & $\mathbf{.3426}\std{.006}$\sig &  \\
\midrule
\multicolumn{10}{l}{\textit{Next visit prediction}} \\
THP~\citep{zuo2020thp} & $.0831\std{.001}$ & $.1036\std{.000}$ & $.0620\std{.003}$ & $.0460\std{.001}$ & $.0610\std{.001}$ & $.0909\std{.002}$ & $.0466\std{.005}$ & $.0742\std{.008}$ & $.1584\std{.004}$ \\
A-NHP~\citep{yang2022anhp} & $.0182\std{.009}$ & $.0219\std{.011}$ & $\underline{.2708}\std{.000}$ & $.0486\std{.003}$ & $.0641\std{.004}$ & $\underline{.2028}\std{.052}$ & $.0528\std{.008}$ & $.0752\std{.008}$ & $\underline{.2496}\std{.002}$ \\
TrajGPT~\citep{hsu2024trajgpt} & $.1106\std{.057}$ & $.1420\std{.070}$ & $.1566\std{.031}$ & $.0475\std{.000}$ & $.0636\std{.002}$ & $.1221\std{.029}$ & $.0562\std{.032}$ & $.0793\std{.041}$ & $.1523\std{.014}$ \\
TAPT~\citep{xu2025tapt} & $\underline{.2745}\std{.006}$ & $\underline{.3690}\std{.008}$ & $.1111\std{.013}$ & $\underline{.2575}\std{.004}$ & $\underline{.3303}\std{.003}$ & $.1044\std{.007}$ & $\underline{.1818}\std{.006}$ & $\underline{.2474}\std{.006}$ & $.1099\std{.004}$ \\
Mobility-LLM~\citep{gong2024mobilityllm} & $.0840\std{.001}$ & $.1063\std{.001}$ & $\mathbf{.4307}\std{.027}$ & $.0536\std{.001}$ & $.0692\std{.001}$ & $\mathbf{.6173}\std{.074}$ & $.0717\std{.006}$ & $.1049\std{.004}$ & $\mathbf{.5614}\std{.073}$ \\
\textbf{\ours{}} & $\mathbf{.3838}\std{.003}$\sig & $\mathbf{.4835}\std{.003}$\sig & $.2181\std{.004}$\sig & $\mathbf{.3203}\std{.007}$\sig & $\mathbf{.3916}\std{.007}$\sig & $.1601\std{.006}$\sig & $\mathbf{.2787}\std{.001}$\sig & $\mathbf{.3465}\std{.003}$\sig & $.1650\std{.011}$\sig \\
\bottomrule
\end{tabular}
\end{table}

\emph{Discussion.}
On next-POI, \ours{} tops H@$10$/H@$20$ on all three cities and every one of those six margins is significant, leading the strongest baseline by $+.050$/$+.088$/$+.072$ H@$10$ (MobTCast on Tokyo, LoTNext on Stockholm/Austin).
On next-visit, \ours{} tops H@$10$/H@$20$ on all three cities as well and every one of those six margins is significant, beating the strongest next-visit baseline TAPT by $+.109$/$+.063$/$+.097$ H@$10$.
Mobility-LLM and A-NHP lead T$\pm60$ on all three cities, at $2.0$--$3.9\times$ and $1.2$--$1.5\times$ our value, respectively, but at $3.9$--$6.0\times$ and $5.3$--$21\times$ lower H@$10$, placing each on a distinct location--time Pareto point.

\subsection{Social link inference}
\label{sec:experiments:social}

Predicting platform-graph links from visit histories alone directly tests A3: co-located visits at overlapping times are the primary observable, and pre-training sees no graph supervision.

\begin{table}[!htbp]
\centering
\scriptsize
\caption{Social link inference results.}
\label{tab:experiments:social}
\setlength{\tabcolsep}{1.5pt}
\begin{tabular}{llllllllll}
\toprule
& \multicolumn{3}{c}{Foursquare-Tokyo} & \multicolumn{3}{c}{Gowalla-Stockholm} & \multicolumn{3}{c}{Gowalla-Austin} \\
\cmidrule(lr){2-4} \cmidrule(lr){5-7} \cmidrule(lr){8-10}
Method & \hc{AP} & \hc{H@10} & \hc{MRR} & \hc{AP} & \hc{H@10} & \hc{MRR} & \hc{AP} & \hc{H@10} & \hc{MRR} \\
\midrule
UniTraj$^{\dagger}$~\citep{zhu2025unitraj} & $.5126\std{.008}$ & $.1570\std{.007}$ & $.0736\std{.002}$ & $.4337\std{.011}$ & $.1273\std{.010}$ & $.0637\std{.003}$ & $.4965\std{.032}$ & $.1549\std{.008}$ & $.0766\std{.004}$ \\
EBM~\citep{pham2013ebm} & $.7317$ & $.3233$ & $.1760$ & $.7445$ & $.4977$ & $.2994$ & $.7845$ & $.3716$ & $.1801$ \\
USRC~\citep{chu2025one} & $.8688\std{.001}$ & $.5347\std{.003}$ & $.2686\std{.002}$ & $.8256\std{.008}$ & $.5282\std{.005}$ & $.2943\std{.005}$ & $\underline{.8983}\std{.009}$ & $.6094\std{.007}$ & $.3352\std{.008}$ \\
H\textsuperscript{3}GNN~\citep{li2025heterogeneous} & $\underline{.9137}\std{.002}$ & $.7080\std{.010}$ & $.4270\std{.011}$ & $\underline{.8625}\std{.003}$ & $.5968\std{.008}$ & $.3658\std{.011}$ & $.8863\std{.003}$ & $.6215\std{.017}$ & $.3377\std{.019}$ \\
LBSN2Vec~\citep{yang2019revisiting} & $.9042\std{.001}$ & $\underline{.7127}\std{.002}$ & $\underline{.4839}\std{.002}$ & $.8208\std{.001}$ & $\underline{.6628}\std{.002}$ & $\underline{.4874}\std{.002}$ & $.8632\std{.002}$ & $\underline{.6220}\std{.003}$ & $\underline{.3970}\std{.002}$ \\
\textbf{\ours{}} & $\mathbf{.9344}\std{.001}$\sig & $\mathbf{.7801}\std{.002}$\sig & $\mathbf{.5347}\std{.011}$\sig & $\mathbf{.9276}\std{.001}$\sig & $\mathbf{.7426}\std{.003}$\sig & $\mathbf{.5870}\std{.005}$\sig & $\mathbf{.9373}\std{.001}$\sig & $\mathbf{.7423}\std{.007}$\sig & $\mathbf{.4997}\std{.024}$\sig \\
\bottomrule
\end{tabular}
\end{table}

\emph{Discussion.}
\ours{} attains the highest mean in all $9$ columns, and every one of them is significant. AP gains are $+.021$/$+.065$/$+.039$ across Tokyo/Stockholm/Austin, and on Gowalla-Austin the margins over the strongest baseline are $+.120$ H@$10$ and $+.103$ MRR.
Clearing every graph-aware baseline (LBSN2Vec, H\textsuperscript{3}GNN, USRC) without any graph supervision at pre-training is direct evidence for A3: co-located visits suffice to recover platform-level social ties.

\subsection{Enterprise authentication logs}
\label{sec:experiments:auth}

We run the same pre-training code, with no architectural changes, on the LANL unified-host-and-network authentication log ($749$ red-team lateral-movement events as ground truth), a non-mobility MESES instance with no native geometry.
We score \ours's pre-trained backbone directly, with no fine-tuning on LANL (scoring protocol in Appendix~\ref{appendix:lanl_scoring}). We report event-level and user-level metrics (per-\texttt{src\_user} max-pool, Tuor convention~\citep{tuor2017deep}).
Appendix~\ref{appendix:lanl_cohort} reports results on two perturbations of the cohort, a different benign-user sample and twice the per-user event cap.

\begin{table}[!htbp]
\centering
\scriptsize
\caption{Anomaly detection on LANL enterprise authentication logs.}
\label{tab:experiments:auth}
\begin{tabular}{lllllll}
\toprule
& \multicolumn{3}{c}{Event-level} & \multicolumn{3}{c}{User-level} \\
\cmidrule(lr){2-4} \cmidrule(lr){5-7}
Method & \hc{AP} & \hc{AUROC} & \hc{Max F1} & \hc{AP} & \hc{AUROC} & \hc{Max F1} \\
\midrule
LogGPT~\citep{han2023loggpt} & $.0055\std{.001}$ & $.7881\std{.008}$ & $.0217\std{.006}$ & $.2172\std{.036}$ & $.8358\std{.011}$ & $.3362\std{.064}$ \\
UniTraj$^{\dagger}$~\citep{zhu2025unitraj} & $.0075\std{.001}$ & $.7998\std{.024}$ & $.0437\std{.001}$ & $.1961\std{.009}$ & $.7936\std{.010}$ & $.2648\std{.014}$ \\
NuLog~\citep{nedelkoski2020nulog} & $.0059\std{.000}$ & $.8558\std{.003}$ & $.0237\std{.002}$ & $.2337\std{.053}$ & $.8359\std{.010}$ & $.3585\std{.059}$ \\
LogBERT~\citep{guo2021logbert} & $.0079\std{.001}$ & $\mathbf{.8877}\std{.006}$ & $.0213\std{.004}$ & $.2311\std{.094}$ & $.8502\std{.008}$ & $.3133\std{.052}$ \\
Tuor-AAAI17~\citep{tuor2017deep} & $\underline{.0112}\std{.003}$ & $\underline{.8680}\std{.008}$ & $\underline{.0604}\std{.019}$ & $.1717\std{.033}$ & $\underline{.8875}\std{.003}$ & $.2875\std{.023}$ \\
DeepLog~\citep{du2017deeplog} & $.0066\std{.002}$ & $.8669\std{.017}$ & $.0238\std{.011}$ & $\underline{.2351}\std{.050}$ & $\mathbf{.8940}\std{.011}$ & $\underline{.3594}\std{.058}$ \\
\textbf{\ours{} (pre-train only)} & $\mathbf{.0173}\std{.011}$ & $.8594\std{.052}$ & $\mathbf{.0683}\std{.030}$ & $\mathbf{.2392}\std{.039}$ & $.8611\std{.033}$ & $\mathbf{.4151}\std{.042}$ \\
\bottomrule
\end{tabular}
\end{table}

\emph{Discussion.}
Every baseline here is a log-anomaly detector trained on LANL with its own self-supervised objective; \ours{} is pre-trained on the same data with its generic objective and scored without fine-tuning, and it lands in the same range as all of them across all six metrics --- attaining the highest mean on four (event-level AP and Max F1, user-level AP and Max F1), with LogBERT leading event-level AUROC and DeepLog user-level AUROC. None of the six differences is significant, so the fair reading is parity rather than a ranking. The MESES recipe therefore carries off mobility without task-specific training: no LANL fine-tuning and no architectural changes.

\subsection{ICU mortality prediction}
\label{sec:experiments:eicu}

In-hospital mortality on two EHR corpora --- eICU-CRD ($197{,}307$ ICU stays, $17{,}673$ deaths, $9.0\%$) and MIMIC-IV ($94{,}407$ stays, $11{,}333$ deaths, $12.0\%$) --- is a second non-mobility MESES instance. Each stay maps into the $(\text{entity},\text{context},\text{time},\text{category})$ schema of \S\ref{sec:method:noise} via $507$ (eICU) and $520$ (MIMIC) clinical-action concepts, with the patient as the entity and the stay as the prediction unit; splits are patient-level so all stays of a patient land in one fold (Appendix~\ref{appendix:split}, fine-tune head in Appendix~\ref{appendix:hyperparameters}). Every sequence baseline with a static-feature path receives the same side features as \ours{}.

\begin{table}[!htbp]
\centering
\scriptsize
\caption{ICU mortality prediction on eICU-CRD and MIMIC-IV.}
\label{tab:experiments:eicu}
\setlength{\tabcolsep}{3.5pt}
\begin{tabular}{lllllllll}
\toprule
& \multicolumn{4}{c}{eICU-CRD} & \multicolumn{4}{c}{MIMIC-IV} \\
\cmidrule(lr){2-5} \cmidrule(lr){6-9}
Method & \hc{AUROC} & \hc{AP} & \hc{Max F1} & \hc{Sens@Sp.9} & \hc{AUROC} & \hc{AP} & \hc{Max F1} & \hc{Sens@Sp.9} \\
\midrule
UniTraj$^{\dagger}$~\citep{zhu2025unitraj} & $.7574\std{.013}$ & $.2993\std{.024}$ & $.3500\std{.016}$ & $.4197\std{.023}$ & $.7681\std{.015}$ & $.3753\std{.029}$ & $.4037\std{.018}$ & $.4271\std{.026}$ \\
Raindrop~\citep{zhang2022graph} & $.8446\std{.004}$ & $.4347\std{.012}$ & $.4434\std{.007}$ & $.5574\std{.010}$ & $.8650\std{.006}$ & $.5820\std{.019}$ & $.5452\std{.013}$ & $.6205\std{.015}$ \\
SimMTM~\citep{dong2023simmtm} & $.8790\std{.002}$ & $.5565\std{.004}$ & $.5162\std{.002}$ & $.6406\std{.005}$ & $.8748\std{.002}$ & $.5725\std{.003}$ & $.5362\std{.004}$ & $.6219\std{.006}$ \\
WaveGNN~\citep{hajisafi2025wavegnn} & $.8086\std{.002}$ & $.3905\std{.004}$ & $.3985\std{.003}$ & $.4815\std{.005}$ & $\underline{.9032}\std{.002}$ & $\underline{.6477}\std{.003}$ & $\underline{.5875}\std{.005}$ & $\underline{.6886}\std{.013}$ \\
STraTS~\citep{tipirneni2022strats} & $\underline{.8938}\std{.001}$ & $\mathbf{.5965}\std{.003}$ & $\underline{.5449}\std{.004}$ & $\underline{.6779}\std{.007}$ & $.8866\std{.001}$ & $.6052\std{.005}$ & $.5546\std{.004}$ & $.6480\std{.004}$ \\
\textbf{\ours{}} & $\mathbf{.9006}\std{.001}$\sig & $\underline{.5705}\std{.004}$\sig & $\mathbf{.5551}\std{.004}$\sig & $\mathbf{.7074}\std{.005}$\sig & $\mathbf{.9149}\std{.002}$\sig & $\mathbf{.6812}\std{.007}$\sig & $\mathbf{.6221}\std{.004}$\sig & $\mathbf{.7432}\std{.006}$\sig \\
\bottomrule
\end{tabular}
\end{table}

\emph{Discussion.}
\ours{} leads $7$ of the $8$ columns, each significantly: all four MIMIC-IV metrics over the strongest EHR baseline there (WaveGNN, by $+.012$ AUROC, $+.034$ AP, $+.035$ Max F1, $+.055$ Sens@Sp.9) and eICU AUROC, Max F1 and Sens@Sp.9 over STraTS.
The single loss is eICU AP, where STraTS's $.5965$ beats our $.5705$ significantly.
That the prototype-conditioned denoising objective clears both corpora's strongest baseline on $7$ of $8$ metrics, on a substrate with no geometry and no mobility, is the transfer claim of \S\ref{sec:axioms} at scale.

\subsection{Ablation study}
\label{sec:experiments:ablation}

We ablate seven design choices (C1--C7) spanning the pre-training objective, prototype routing, and backbone, with each variant pre-trained and fine-tuned on UA-Berlin (mobility AD), Gowalla-Austin (next-visit, social link), and LANL (auth-log AD). All rows of a column family use the same fixed co-occurrence route. The protocol differs from that of the per-task tables (Appendix~\ref{appendix:ablation_protocol}), so the Full \ours{} row of Table~\ref{tab:experiments:ablation} does not match those tables. Cells with $\pm$ report the mean and standard deviation over five seeds; cells without $\pm$ are single runs.
\begin{table}[!htbp]
\centering
\scriptsize
\caption{Ablation results.}
\label{tab:experiments:ablation}
\setlength{\tabcolsep}{1.75pt}
\begin{tabular}{lllllllll}
\toprule
& \multicolumn{2}{c}{Mobility AD} & \multicolumn{2}{c}{Next-Visit} & \multicolumn{2}{c}{Social Link} & \multicolumn{2}{c}{Auth-Log AD} \\
\cmidrule(lr){2-3} \cmidrule(lr){4-5} \cmidrule(lr){6-7} \cmidrule(lr){8-9}
Variant & \hc{AP} & \hc{MaxF1} & \hc{H@10} & \hc{T$\pm$60} & \hc{AP} & \hc{H@10} & \hc{AP} & \hc{MaxF1} \\
\midrule
\textbf{Full \ours} & $.4290\std{.018}$ & $.4386\std{.021}$ & $.2800\std{.003}$ & $.1641\std{.018}$ & $.9009\std{.006}$ & $.6495\std{.006}$ & $.2392\std{.039}$ & $.4151\std{.042}$ \\
\midrule
C1: w/o $\mathcal{L}_{\text{prototype}}$ & $.1249$ & $.2162$ & $.1471$ & $.1057$ & $.8893\std{.010}$ & $.6489\std{.016}$ & $.1440$ & $.2812$ \\
C2: w/o $\mathcal{L}_{\text{noise}}$ & $.3855$ & $.3729$ & $.2588$ & $.1510$ & $.8844\std{.011}$ & $.5895\std{.052}$ & $.0550$ & $.1165$ \\
C3: w/o prototype-as-input & $.1671$ & $.2235$ & $.1787$ & $.1190$ & $.8826\std{.012}$ & $.6368\std{.020}$ & $.1621$ & $.3158$ \\
C4: next-token pretrain & $.3629$ & $.4046$ & $.2810\std{.002}$ & $.1639\std{.011}$ & $.9091\std{.003}$ & $.6765\std{.010}$ & $.1584$ & $.2222$ \\
C5: masked-token pretrain & $.4066\std{.022}$ & $.4040\std{.018}$ & $.2893\std{.004}$ & $.1678\std{.007}$ & $.9090\std{.003}$ & $.6739\std{.005}$ & $.1499$ & $.2785$ \\
C6: no pre-training & $.1320$ & $.2161$ & $.2587$ & $.1522$ & $.9170\std{.003}$ & $.6907\std{.009}$ & $.2778\std{.097}$ & $.3823\std{.081}$ \\
C7: flat transformer encoder & $.1241$ & $.2192$ & $.0410$ & $.1318$ & $.8965\std{.006}$ & $.6425\std{.016}$ & $.0186$ & $.0479$ \\
\bottomrule
\end{tabular}
\end{table}

\emph{Discussion.}
C1 (no $\mathcal{L}_{\text{prototype}}$) drops mobility-AD AP by $71\%$ and next-visit H@10 by $47\%$, and C3 (prototype kept as a contrastive positive but removed from input tokens) is the weakest arm on social-link AP. C1 and C3 score below Full in every column.
The arms that \emph{substitute} one pre-training objective for another --- C4 (autoregressive next-token) and C5 (masked-token, UniTraj-style~\citep{zhu2025unitraj}) --- do not behave uniformly across tasks: Full leads C5 by $+.035$ Max F1 on mobility anomaly detection, whereas C5 leads Full by $+.009$ next-visit H@10, and C4 and C5 lead Full on social link by $+.008$ AP and by $+.027$ and $+.024$ H@10. Each of these differences is significant ($p \leq .03$, one-sided Wilcoxon rank-sum over five seeds). C6 (no pre-training) is the strongest social-link arm. On next-visit T$\pm60$, C4 and C5 are within $.004$ of Full, and C1 and C3 are $.058$ and $.045$ below it.
C7 (flat transformer) is the weakest variant on next-visit H@10 ($.0410$ against Full's $.2800$) and on both auth-log AD metrics ($.0186$/$.0479$ against $.2392$/$.4151$); on social link it trails Full by under $.005$ AP and $.007$ H@10, and neither difference is significant ($p \geq .15$).
Across tasks, the entity prototype (C1, C3) and the factorized attention (C7) account for the largest drops on anomaly detection and next-visit prediction, the denoising objective (C2, C5) helps mainly on anomaly detection, and on social link the strongest arm is the one without pre-training (C6).

\section{Conclusion}
\label{sec:conclusion}

We introduced \emph{multi-entity spatiotemporal event streams} (MESES), a class of data found in domains as varied as mobility, security, and clinical events. We formalized three axioms (A1--A3) that any pre-training framework for MESES should satisfy, and showed that the dominant next-token pre-training recipes --- autoregressive and masked alike --- enforce none of the three. 

We present \ours{} as a framework that satisfies all three. \ours's noise-detection and entity-prototype objectives on a co-occurrence-augmented backbone are jointly designed to satisfy A1--A3, and a single \ours{} checkpoint per dataset matches or beats specialized baselines across six mobility cohorts and four tasks.
Ablations are consistent with the axiom-aligned mechanism where the axioms bite hardest: removing the prototype objective collapses anomaly detection ($-71\%$ AP) and next-visit ranking ($-47\%$ H@$10$), and removing its input-side role gives the largest social-link AP drop.
The same backbone transfers without architectural change to two non-mobility instances: scored without fine-tuning on the LANL authentication log, it is statistically indistinguishable from five dedicated log-anomaly detectors, and on eICU-CRD and MIMIC-IV it leads seven of eight mortality-prediction columns. That is what makes \ours{} a framework for MESES rather than a domain-specific tool.

The MESES framing extends in two ways.
Many other event streams (e-commerce sessions, telecom usage, IoT readings, public-health surveillance) fit the same $(\text{entity}, \text{context}, \text{time}, \text{category})$ schema, and pre-training a unified \ours{} backbone over a multi-instance MESES corpus is the natural next step.
More broadly, \ours{} illustrates a principle that should generalize beyond MESES: pre-training objectives are most effective when designed to match structural properties of the data, rather than imported from other domains.
Identifying analogous axioms for other data classes (multi-modal sensor streams, knowledge graphs, scientific time series) is a natural extension of this principled methodology.

\paragraph{Limitations.}
\ours{} is discriminative, not a TPP substitute (it trails on T$\pm60$); the co-occurrence sub-layer is dispatched per seed on a validation criterion rather than learned; and we pre-train one checkpoint per dataset rather than a unified MESES backbone (Appendix~\ref{appendix:limitations}).

\begin{ack}
This research has been funded in part by the NSF award DMS-2428039, and the Intelligence Advanced Research Projects Activity (IARPA) via the Department of Interior/Interior Business Center (DOI/IBC) contract number 140D0423C0033. The U.S. Government is authorized to reproduce and distribute reprints for Governmental purposes, notwithstanding any copyright annotation thereon. Disclaimer: The views and conclusions contained herein are those of the authors and should not be interpreted as necessarily representing the official policies or endorsements, either expressed or implied, of IARPA or the U.S. Government. The authors gratefully acknowledge the compute and data resources provided by Google Cloud Platform through the Google Cloud Research Credits program, as well as the unrestricted cash award from Google Research.
\end{ack}

\bibliographystyle{plainnat} % NeurIPS requires a natbib-compatible style
\bibliography{reference}    % Points to reference.bib (do not include the .bib extension)

\appendix
\section{Formal description of \ours}
\label{appendix:formal}

This section restates the architecture and pre-training losses of \S\ref{sec:method} in self-contained mathematical form.
All notation is inherited from \S\ref{sec:meses} and \S\ref{sec:method}; symbols introduced here are local to the appendix and italicized in the index below.
Figure~\ref{fig:traxion_arch} in \S\ref{sec:method:arch} gives a schematic of the full pre-training pipeline; this section gives the same content in equations.

\paragraph{Notation.}
We write $u \in \mathcal{U}$ for an entity, $\mathcal{X}$ for the substrate with embedding $\phi : \mathcal{X} \to \mathbb{R}^{d_x}$, and
$e_{u,t} = (u, x_{u,t}, \tau_{u,t}, a_{u,t}, \delta_{u,t})$ for the $t$-th event of $u$, with $a_{u,t} \in \Delta^{n_a-1}$ a probability vector over $n_a$ activity categories and $\delta_{u,t} \in \mathbb{R}_{\geq 0} \cup \{\bot\}$ an optional duration.
A pre-training example consists of a chronologically ordered window
$E_u = (e_{u,1}, \ldots, e_{u,T})$ of length $T$ and, for each event $t$, a tuple of co-occuring peers from other entities (defined below).
Let $d$ denote the model dimension and $F$ the number of feature tokens per event, with $d_f = d/F$ the per-token width.
The encoder $f_\theta$ produces contextualized embeddings $\mathcal{H}_{u,t} \in \mathbb{R}^d$ as in Eq.~\ref{eq:loss};
the prototype table is $P \in \mathbb{R}^{|\mathcal{U}| \times d_f}$ with rows $p_u$.

\subsection{Per-event feature embedding}
\label{appendix:formal:embed}

Each event $e_{u,t}$ is mapped to feature tokens
$\bigl(\mathbf{z}^{\mathrm{loc}}_{u,t},\, \mathbf{z}^{\mathrm{start}}_{u,t},\, \mathbf{z}^{\mathrm{stop}}_{u,t},\, \mathbf{z}^{\mathrm{ent}}_{u,t},\, \mathbf{z}^{\mathrm{act}}_{u,t}\bigr) \in (\mathbb{R}^{d_f})^F$
that are stacked along a feature axis.
We use $F = 5$ on every instance except the two EHR corpora, where the backbone width $d{=}128$ (with $H{=}4$ heads and $2$ layers) is not divisible by $5$;
on those two we drop $\mathbf{z}^{\mathrm{ent}}_{u,t}$ from the input stack, leaving $F = 4$.
The prototype $p_u$ is still drawn from the same table $P$ and used as the supervisory target in Eq.~\ref{eq:prototype}, so the contrastive coupling between the encoder and entity identity is preserved; the consequence is that A2's input-side entity-feature clause (\S\ref{sec:axioms}) is unmet on the two EHR corpora while the supervisory-side clause holds.
The smallest width that retains the entity token at the same head count is $d{=}160$ ($\sim 25\%$ more parameters at the same depth), which was not run within our compute envelope; we record this as a follow-up in Appendix~\ref{appendix:limitations}.
Point events ($\delta_{u,t} = \bot$) keep both time tokens with $\mathbf{z}^{\mathrm{stop}}_{u,t} = \mathbf{z}^{\mathrm{start}}_{u,t}$ rather than collapsing to a single token.

\paragraph{Spatial token (Space2Vec).}
Let $\{a_1, a_2, a_3\}$ be the unit vectors of a planar hexagonal lattice with $a_1 = (1,0)$, $a_2 = (-\tfrac12, \tfrac{\sqrt 3}{2})$, $a_3 = (-\tfrac12, -\tfrac{\sqrt 3}{2})$, and let
\begin{equation}
\lambda_s = \lambda_{\min} \, \Bigl(\tfrac{\lambda_{\max}}{\lambda_{\min}}\Bigr)^{s/(N_s-1)}, \qquad s = 0, \ldots, N_s-1,
\label{eq:space2vec_scales}
\end{equation}
be a geometric sweep of spatial scales over $[\lambda_{\min}, \lambda_{\max}]$.
Define the multi-scale phase
$\rho_{j,s}(x) = \langle a_j, \phi(x) \rangle / \lambda_s$ for $j \in \{1,2,3\}$.
Stacking the $6 N_s$ values $(\cos \rho_{j,s}, \sin \rho_{j,s})$ into a vector $\mathrm{PE}(x) \in \mathbb{R}^{6 N_s}$ and applying a learned linear-ReLU map gives
\begin{equation}
\mathbf{z}^{\mathrm{loc}}_{u,t} \;=\; \mathrm{ReLU}\bigl(W_{\mathrm{loc}} \, \mathrm{PE}(x_{u,t})\bigr) \in \mathbb{R}^{d_f},
\qquad W_{\mathrm{loc}} \in \mathbb{R}^{d_f \times 6 N_s}.
\label{eq:space2vec}
\end{equation}

\paragraph{Temporal tokens (Time2Vec).}
Let $\Pi : \mathbb{R} \to [0, P)$ denote modular wrapping by the dataset-specific period $P$ (\texttt{daily}, \texttt{weekly}, or \texttt{none}; Appendix~\ref{appendix:design_intuitions}).
With learned $w \in \mathbb{R}^{d_f}$, $b \in \mathbb{R}^{d_f}$, the Time2Vec encoding is the channel-wise map
\begin{equation}
\mathrm{T2V}(\tau)_k =
\begin{cases}
w_0 \, \Pi(\tau) + b_0, & k = 0, \\
\sin\!\bigl(w_k \, \Pi(\tau) + b_k\bigr), & k = 1, \ldots, d_f - 1,
\end{cases}
\label{eq:time2vec}
\end{equation}
giving $\mathbf{z}^{\mathrm{start}}_{u,t} = \mathrm{T2V}(\tau_{u,t})$ and $\mathbf{z}^{\mathrm{stop}}_{u,t} = \mathrm{T2V}(\tau_{u,t} + \delta_{u,t})$, with the convention $\delta_{u,t}{=}0$ when $\delta_{u,t}=\bot$ (point events have $\mathbf{z}^{\mathrm{stop}}_{u,t} = \mathbf{z}^{\mathrm{start}}_{u,t}$).

\paragraph{Entity token (factored prototype).}
The prototype table is factored as $p_u = W_P \, q_u$ with a small lookup $q_u \in \mathbb{R}^{h}$ and a shared projection $W_P \in \mathbb{R}^{d_f \times h}$, so that
\begin{equation}
\mathbf{z}^{\mathrm{ent}}_{u,t} \;=\; p_u \;=\; W_P \, q_u \;\in\; \mathbb{R}^{d_f}.
\label{eq:prototype_lookup}
\end{equation}
The same $p_u$ is used as input feature here and as supervisory target in Eq.~\ref{eq:prototype} (\S\ref{sec:method:user_loss}); only $\{q_u\}_{u \in \mathcal{U}}$ and $W_P$ are stored.
Setting $h \ll d_f$ controls the parameter cost when $|\mathcal{U}|$ is large.

\paragraph{Activity token.}
With $W_a \in \mathbb{R}^{d_f \times n_a}$,
\begin{equation}
\mathbf{z}^{\mathrm{act}}_{u,t} \;=\; W_a \, a_{u,t} \;\in\; \mathbb{R}^{d_f}.
\label{eq:activity}
\end{equation}

\paragraph{Per-event matrix.}
We collect the $F$ tokens into a per-event matrix
$Z_{u,t} = [\mathbf{z}^{\mathrm{loc}}_{u,t}, \mathbf{z}^{\mathrm{start}}_{u,t}, \mathbf{z}^{\mathrm{stop}}_{u,t}, \mathbf{z}^{\mathrm{ent}}_{u,t}, \mathbf{z}^{\mathrm{act}}_{u,t}]^\top \in \mathbb{R}^{F \times d_f}$
and write $\mathbf{Z}_u = (Z_{u,1}, \ldots, Z_{u,T}) \in \mathbb{R}^{T \times F \times d_f}$ for the window.

\subsection{Co-occurrence retrieval}
\label{appendix:formal:coloc}

For each focal event $e_{u,t}$ we form a peer set
\begin{equation}
\mathcal{N}_{u,t} \;=\; \mathrm{top}^{\,(C-1)}_{e \in \mathcal{P}(u,t)}\;\bigl|\tau - \tau_{u,t}\bigr| + \bigl|(\tau + \delta) - (\tau_{u,t} + \delta_{u,t})\bigr|,
\label{eq:peer_topk}
\end{equation}
where $\mathcal{P}(u,t) = \{e_{u',t'} : x_{u',t'} = x_{u,t},\; u' \neq u\}$ is the set of training events from \emph{other} entities at the same substrate context (taking $\delta = 0$ when $\delta = \bot$), $\mathrm{top}^{\,(C-1)}$ keeps the $C-1$ smallest values, and $\mathcal{N}_{u,t}$ is padded to size $C-1$ with masked dummies when fewer than $C-1$ candidates exist (Appendix~\ref{appendix:co-location}).
Let $Z^{(c)}_{u,t}$ be the per-event feature matrix of the $c$-th peer for $c = 1, \ldots, C-1$, and write the focal slot as $Z^{(0)}_{u,t} \!=\! Z_{u,t}$.
Stacking peers along a co-occurrence axis yields the rank-5 tensor
\begin{equation}
X^{(0)}_u \;=\; \bigl[\,Z^{(0)}_{u,t}, Z^{(1)}_{u,t}, \ldots, Z^{(C-1)}_{u,t}\,\bigr]_{t=1}^{T} \;\in\; \mathbb{R}^{T \times F \times C \times d_f},
\label{eq:rank5}
\end{equation}
together with a peer mask $M_u \in \{0,1\}^{T \times C}$ ($M_{u,t,0} = 0$ always; $M_{u,t,c} = 1$ for padded slots) and a sequence padding mask $m_u \in \{0,1\}^{T}$ from window padding.

\subsection{Factorized attention block}
\label{appendix:formal:attn}

Let $\mathrm{TF}^{(\mathrm{ax})}_{H,d_{\mathrm{ff}}}$ denote a standard pre-LN Transformer encoder layer (multi-head attention with $H$ heads followed by a position-wise MLP of width $d_{\mathrm{ff}}$, with residual connections and LayerNorm) applied to a tensor by treating axis $\mathrm{ax}$ as the token axis and all remaining non-embedding axes as independent batch axes.
The \ours backbone stacks $L$ blocks $X^{(\ell+1)}_u = \mathrm{Block}_{\ell+1}(X^{(\ell)}_u)$, $\ell = 0, \ldots, L-1$, where each block consists of three axis-restricted sub-layers acting on $X \in \mathbb{R}^{T \times F \times C \times d_f}$:
\begin{align}
Y &\;=\; \mathrm{TF}^{(F)}_{H, d}(X), &&\text{(feature-axis attention; all $T \cdot C$ slots)}\label{eq:tf_feat}\\
\widetilde{Y}_{t, f, 0,:} &\;=\; \mathrm{XAttn}\!\bigl(\,Y_{t,f,0,:};\; \{Y_{t,f,c,:}\}_{c=0}^{C-1};\; M_{u,t,:}\,\bigr),
&&\text{(focal $\to$ peer cross-attention)}\label{eq:tf_neigh}\\
\widetilde{Y}_{t, f, c,:} &\;=\; Y_{t, f, c,:} \quad (c \geq 1), && \notag\\
X^{\prime}_{:, :, 0, :} &\;=\; \mathrm{TF}^{(T)}_{H, d}\!\bigl(\widetilde{Y}_{:, :, 0, :};\; m_u\bigr), &&\text{(sequence-axis attention; focal slot only)}\label{eq:tf_seq}\\
X^{\prime}_{:, :, c, :} &\;=\; \widetilde{Y}_{:, :, c, :} \quad (c \geq 1). && \notag
\end{align}
$\mathrm{XAttn}(q;\, K;\, M)$ is a single multi-head cross-attention sub-layer (with residual, LayerNorm, and position-wise MLP) where $q$ is the focal query, $\{K_c\}$ are the peer keys/values, and $M$ is the key-padding mask;
the asymmetric query-only update rewrites only the focal slot's representation in this sub-layer (peer slots retain $Y_{t,f,c,:}$ for $c \geq 1$); each peer event is itself the focal slot in its own training window, so a symmetric self-attention here would be redundant and $C\times$ more expensive.

After $L$ blocks, the contextualized event embedding is the per-event flattening of the focal slot's $F$ tokens:
\begin{equation}
\mathcal{H}_{u,t} \;=\; \mathrm{vec}\bigl(X^{(L)}_{u,t,:,0,:}\bigr) \;\in\; \mathbb{R}^{d}, \qquad H_u = (\mathcal{H}_{u,1}, \ldots, \mathcal{H}_{u,T}) \in \mathbb{R}^{T \times d}.
\label{eq:focal_readout}
\end{equation}
This $\mathcal{H}_{u,t}$ is the same vector referenced by Eq.~\ref{eq:loss}--\ref{eq:prototype}.

\subsection{Pre-training heads and the perturbation operator}
\label{appendix:formal:heads}

\paragraph{Noise-detection head.}
A linear head $g_\psi : \mathbb{R}^d \to \mathbb{R}$ scores each event for being perturbed:
\begin{equation}
g_\psi(\mathcal{H}_{u,t}) \;=\; w_\psi^\top \mathcal{H}_{u,t} + b_\psi.
\label{eq:bce_head}
\end{equation}
Eq.~\ref{eq:denoise} is then the masked binary cross-entropy of $g_\psi(\mathcal{H}_{u,t})$ against the per-event target $y_{u,t}$ produced by $\eta$, averaged over the non-padded events $\mathcal{V} = \{(u,t) : m_{u,t} = 0\}$.

\paragraph{Prototype-contrastive projection.}
The InfoNCE anchor $\bar{\mathcal{H}}_{u,t}$ in Eq.~\ref{eq:prototype} is obtained by passing $\mathcal{H}_{u,t}$ through a two-layer MLP $\pi_\theta : \mathbb{R}^d \to \mathbb{R}^{d_f}$ and $\ell_2$-normalizing,
\begin{equation}
\pi_\theta(h) \;=\; W_2\,\mathrm{ReLU}(W_1 h + b_1) + b_2,
\qquad
\bar{\mathcal{H}}_{u,t} \;=\; \frac{\pi_\theta(\mathcal{H}_{u,t})}{\|\pi_\theta(\mathcal{H}_{u,t})\|_2},
\qquad
\bar{p}_u \;=\; \frac{p_u}{\|p_u\|_2},
\label{eq:proj_norm}
\end{equation}
with $W_1 \in \mathbb{R}^{h_{\mathrm{proj}} \times d}$, $W_2 \in \mathbb{R}^{d_f \times h_{\mathrm{proj}}}$.
Anchors are restricted to $\mathcal{V}^+ = \{(u,t) \in \mathcal{V} : y_{u,t} = 0\}$ (Appendix~\ref{appendix:anchoring});
the denominator of Eq.~\ref{eq:prototype} ranges over the prototypes of all minibatch entities $B \subseteq \mathcal{U}$ that have at least one anchor.

\paragraph{Perturbation operator.}
The operator $\eta$ of \S\ref{sec:method:noise} is a stochastic mapping $\eta : \mathcal{E}^T \to \mathcal{E}^T \times \{0,1\}^T$ on event windows.
Let $\mathrm{Bern}(p)$ denote a Bernoulli draw and $\mathrm{Unif}$ a uniform draw on the indicated set.
Sample $b \sim \mathrm{Bern}(1 - p_{\mathrm{norm}})$;
if $b = 0$ return $(E_u, \mathbf{0}_T)$.
Otherwise draw an indicator vector $y_u \in \{0,1\}^T$ with $y_{u,t} \stackrel{\mathrm{iid}}{\sim} \mathrm{Bern}(r)$ at a fixed rate $r = 0.3$, forcing $\sum_t y_{u,t} \geq 1$.
For each flagged $t$ (i.e.\ $y_{u,t} = 1$), draw $\kappa_{u,t} \sim \mathrm{Unif}\{\text{loc}, \text{time}, \text{both}\}$ and apply
\begin{equation}
\eta_{\mathrm{loc}}(e_{u,t}): \;
x_{u,t}' \;=\; \arg\min_{x \in \mathcal{X}}\bigl\|\phi(x) - \tilde{\phi}\bigr\|_2,
\quad \tilde{\phi} \sim \mathrm{Unif}(\mathrm{AOI}),
\quad a_{u,t}' = a(x_{u,t}'),
\label{eq:eta_loc}
\end{equation}
\begin{equation}
\eta_{\mathrm{time}}(e_{u,t}): \;
\tau_{u,t}' \;\sim\; \mathrm{Unif}\!\bigl(\tau_{u,t-1}+\delta_{u,t-1},\; \tau_{u,t+1}\bigr),
\quad x_{u,t}' = x_{u,t},
\label{eq:eta_time}
\end{equation}
applied separately or in composition $\eta_{\mathrm{loc}} \circ \eta_{\mathrm{time}}$ depending on $\kappa_{u,t}$.
When the interval in Eq.~\ref{eq:eta_time} is empty --- e.g.\ concurrent events with $\tau_{u,t-1}+\delta_{u,t-1} > \tau_{u,t+1}$, or boundary positions $t \in \{1, T\}$ where a neighbor is undefined --- $\eta_{\mathrm{time}}$ falls back to the identity on that event and $y_{u,t}$ is set to $0$.
$\mathrm{AOI} \subset \mathbb{R}^{d_x}$ is the dataset's area-of-interest in $\phi$-space, and $a(x)$ denotes the activity associated with substrate context $x$.
The output of $\eta$ is the corrupted window $\tilde E_u$ and label vector $y_u$ used in Eq.~\ref{eq:denoise}.

\subsection{Total objective}
\label{appendix:formal:loss}

For a minibatch of windows $\{E_u\}_{u \in B}$, sample $(\tilde E_u, y_u) = \eta(E_u)$ independently per entity and compute $\{\mathcal{H}_{u,t}\}$ via Eq.~\ref{eq:focal_readout}.
With Eq.~\ref{eq:denoise} and Eq.~\ref{eq:prototype} the joint pre-training loss of Eq.~\ref{eq:loss} reads
\begin{equation}
\mathcal{L}(\theta, \psi)
\;=\;
\underbrace{\frac{1}{|\mathcal{V}|}\!\!\sum_{(u,t) \in \mathcal{V}}\!\!\mathrm{BCE}\!\bigl(g_\psi(\mathcal{H}_{u,t}),\, y_{u,t}\bigr)}_{\mathcal{L}_{\text{noise}}(\theta, \psi)}
\;+\;
\gamma \cdot
\underbrace{\frac{1}{|\mathcal{V}^+|}\!\!\!\!\sum_{(u,t) \in \mathcal{V}^+}\!\!\!\!
- \log\!\frac{\exp\!\bigl(\bar{\mathcal{H}}_{u,t}^\top \bar p_u / \beta\bigr)}{\sum_{u' \in B}\exp\!\bigl(\bar{\mathcal{H}}_{u,t}^\top \bar p_{u'} / \beta\bigr)}}_{\mathcal{L}_{\text{prototype}}(\theta)}.
\label{eq:loss_full}
\end{equation}
Optimization uses AdamW with cosine decay; concrete values of $T, F, C, L, H, d, d_f, h, h_{\mathrm{proj}}, N_s, [\lambda_{\min}, \lambda_{\max}], P, \lambda, \beta, p_{\mathrm{norm}}$ are listed in Appendix~\ref{appendix:hyperparameters}.

\section{Implementation details of \ours}
\label{appendix:implementation}

\subsection{Pre-training mechanics}
\label{appendix:pretraining}

\subsubsection{Selection and composition of pre-training perturbation}
\label{appendix:perturb_composition}
Within a corrupted window we flag a random proportion of events for perturbation; each flagged event is then assigned context perturbation, time perturbation, or both, with equal probability.
If no event ends up flagged (for example, because rejection sampling fails on a short window), we force at least one so that the denoising objective is non-trivial on every corrupted example.
If both perturbation types are disabled the operator degenerates to the identity and Eq.~\ref{eq:denoise} is masked out.

\subsubsection{Anchoring the prototype loss on unperturbed events only}
\label{appendix:anchoring}
Negative examples in Eq.~\ref{eq:prototype} are drawn from minibatch entities, but anchors --- the event representations $\bar{\mathcal{H}}_{u,t}$ that appear in the InfoNCE numerator --- are restricted to the events the denoising operator $\eta$ left unchanged ($\mathcal{V}^+$).
A corrupted anchor would pull $\bar{p}_u$ toward an event the entity did not actually produce, blurring the prototype's role as a summary of the entity's normal behavior.
Restricting the anchor set is what keeps the two pre-training objectives compatible: the denoising objective uses corrupted events as positive supervision for detecting perturbations, while the prototype objective uses unperturbed events as positive supervision for binding events to their entities.

\subsubsection{Why the shared-prototype design does not collapse to an identity shortcut}
\label{appendix:identity_shortcut}
With $p_u$ as an input token and residual connections in the backbone, a trivial identity path could drive Eq.~\ref{eq:prototype} to zero; joint training with $\mathcal{L}_{\text{noise}}$ (Eq.~\ref{eq:denoise}) prevents this by forcing the encoder to model the joint context/time/activity distribution rather than copy $p_u$.

\subsubsection{Co-occurrence retrieval}
\label{appendix:co-location}
The co-occurrence axis is assembled by a non-parametric lookup over a context-keyed inverted index of substrate events.
At load time, for each focal event $e_{u,t}$ we query the index for co-occurring events at context $x_{u,t}$ (excluding events of the focal entity $u$), optionally filter to those whose $[\tau,\, \tau+\delta]$ interval overlaps $[\tau_{u,t},\, \tau_{u,t}+\delta_{u,t}]$ by at least a threshold fraction, and rank the remainder by the Eq.~\ref{eq:peer_topk} distance; the top $C{-}1$ are kept and the rest are padded and masked out of attention.
Each retrieved row contributes the substrate tuple of \S\ref{sec:meses} into the cross-attention as keys/values, so the focal stream sees only event-tuple features from peers; the index is built once per data partition (separately for train, val, and test; a one-shot precompute) and the padding mask is carried through every co-occurrence sub-layer so events with no co-occurring peers are handled gracefully.

\paragraph{Cost and scaling of the retrieval.}
The inverted index is a hash table mapping each context $x \in \mathcal{X}$ to the row positions of the training events at $x$, built in one linear pass over the partition with $\mathcal{O}(N)$ time and $\mathcal{O}(N)$ memory ($4$ bytes per event), where $N$ denotes the number of training events; on NUMOSIM-LA ($N \approx 3.4 \times 10^{7}$ events over $|\mathcal{X}| = 920{,}360$ contexts; Table~\ref{tab:datasets}) the index itself occupies $\sim$$140$\,MB and is paid once at dataset construction.
A single peer query for focal event $e_{u,t}$ costs an $\mathcal{O}(1)$ hash lookup of the bucket at $x_{u,t}$, an $\mathcal{O}(k_x)$ self-exclusion and optional overlap filter, and an $\mathcal{O}(k_x \log k_x)$ partial sort over the bucket to select the $C{-}1$ peers minimizing the Eq.~\ref{eq:peer_topk} distance, where $k_x$ denotes the per-bucket fan-in (the number of training events at context $x$).
Per-query cost therefore scales with the per-bucket fan-in $k_x$: the average fan-in $N / |\mathcal{X}|$ across our datasets is $\approx 37$ (NUMOSIM-LA), $\approx 10$ (Foursquare-Tokyo), and $\approx 86$ (LANL-auth).
Retrieval runs inside CPU $\mathrm{DataLoader}$ workers in parallel with GPU forward/backward, keeping the lookup off the model's critical path at our settings; at far larger scale ($N = 10^{9}$ events, $|\mathcal{X}| = 10^{7}$ contexts) the index reaches $\sim$$4$\,GB and per-query cost continues to scale with bucket fan-in, and standard mitigations (sharding $\mathcal{X}$ across workers, replacing the partial sort with $\mathcal{O}(k_x)$ partitioning, or capping each bucket to a fixed reservoir) preserve the structure of Eq.~\ref{eq:peer_topk}.

\subsubsection{Design rationale}
\label{appendix:design_intuitions}

\paragraph{Location and temporal perturbations.}
Before encoding, we wrap the raw timestamp $t$ (hours from the dataset origin) by a per-dataset period: $t \gets t \bmod 24$, $t \gets t \bmod 168$, or no wrapping, selected by validation loss (Table~\ref{tab:hparam_pretrain}).
Location perturbation dominates the pre-training denoising signal because it samples city-scale coordinates, producing a much larger Euclidean shift in Space2Vec space than time perturbation's feasible-window resample.

\begin{figure}[!htbp]
\centering
\includegraphics[width=0.55\linewidth]{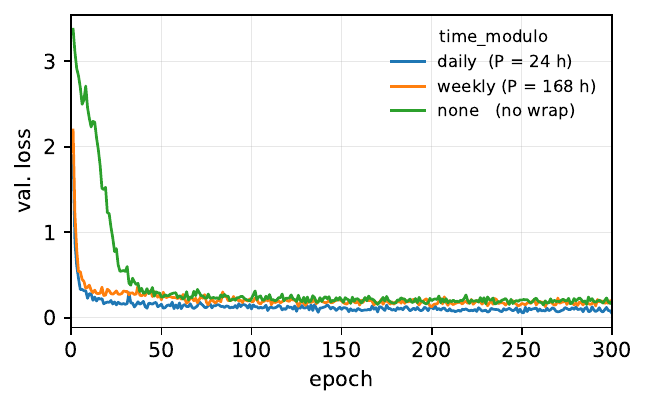}
\caption{UA-Berlin pre-training validation loss under three wrap settings (identical recipe, $300$ epochs).  Wrapping at $24\,\mathrm{h}$ converges to roughly $4{\times}$ lower loss than wrapping at $168\,\mathrm{h}$ or no wrapping on this dataset; the wrap period is a per-dataset hyperparameter.}
\label{fig:time_modulo_val_loss}
\end{figure}

\paragraph{Empirical validation.}
In a matched UA-Berlin pre-train (identical recipe, $300$ epochs), final validation total-loss was $0.052$ at $\bmod\,24$, $0.201$ at $\bmod\,168$, and $0.192$ without wrapping (Figure~\ref{fig:time_modulo_val_loss}); the per-dataset choices in Table~\ref{tab:hparam_pretrain} come from analogous sweeps.

\subsection{Hyperparameters}
\label{appendix:hyperparameters}

We report the settings used to produce Tables~\ref{tab:experiments:humob-anomaly}, \ref{tab:experiments:next_visit}, and~\ref{tab:experiments:social}.
All runs share the architecture and loss defaults below, except for the backbone width and depth of the two ICU corpora; the dataset- and task-specific settings are listed in Tables~\ref{tab:hparam_pretrain} and~\ref{tab:hparam_finetune}.

\paragraph{Shared architecture and loss.}
We set $\gamma$ in Eq.~\ref{eq:loss} to $0.5$, $p_{\text{norm}}=0.7$, $\beta$ in Eq.~\ref{eq:prototype} to $0.07$, $N_{s}$ in \S\ref{sec:method:feature} to $32$, and $h$ in \S\ref{sec:method:agent} to $32$.
The factorized attention in \S\ref{sec:method:attn} stacks $L=6$ blocks with $H=4$ heads and a feed-forward width of $1024$; we use $d=1040$ (auto-adjusted from a target of $1024$ so that $d/F$ is integral). The two ICU corpora use $d=128$ and $L=2$ (Appendix~\ref{appendix:formal:embed}).
Sequences are chunked into non-overlapping windows of $T=32$ events; co-occurrence retrieval uses no minimum overlap filter. The two ICU corpora are the one exception: they pre-train at $T=32$ but fine-tune at $T=128$, because the mortality label is defined over a whole stay rather than a window of it.

\paragraph{Per-dataset pre-training settings.}
Pre-training uses AdamW~\citep{loshchilov2019adamw} with peak learning rate $2{\times}10^{-4}$, gradient-norm clipping at $1.0$, and a cosine schedule decaying to $\eta_{\min}{=}10^{-6}$ over the full budget.
Early stopping monitors an EMA-smoothed total validation loss (EMA factor $0.1$).
Table~\ref{tab:hparam_pretrain} lists the dataset-varying settings: co-occurrence window size $C$, Space2Vec scale range $[\lambda_{\min},\lambda_{\max}]$, the wrap period applied to timestamps before Time2Vec~\citep{kazemi2019time2vec} (selected by validation loss), weight decay, maximum epoch budget, and patience.

\begin{table}[!htbp]
\centering
\scriptsize
\setlength{\tabcolsep}{4pt}
\begin{tabular}{lcccccc}
\toprule
Dataset & $C$ & $[\lambda_{\min},\lambda_{\max}]$ & modulo & weight decay & max epochs & patience \\
\midrule
NUMOSIM-LA          & 8 & $[10^{-6},\,2]$     & weekly & $10^{-3}$ & 200 & 20 \\
UA-Atlanta          & 8 & $[10^{-6},\,2]$     & daily  & $10^{-3}$ & 200 & 40 \\
UA-Berlin           & 4 & $[10^{-6},\,2]$     & daily  & $10^{-2}$ & 800 & 80 \\
Foursquare-Tokyo    & 8 & $[10^{-3},\,360]$   & daily  & $10^{-2}$ & 400 & 40 \\
Gowalla-Stockholm   & 4 & $[10^{-3},\,360]$   & daily  & $10^{-2}$ & 500 & 80 \\
Gowalla-Austin      & 4 & $[10^{-3},\,360]$   & daily  & $10^{-2}$ & 500 & 80 \\
LANL                & 8 & $[10^{-3},\,2]$     & daily  & $10^{-2}$ & 200 & 40 \\
eICU-CRD            & 4 & $[10^{-6},\,2]$     & daily  & $10^{-3}$ & 80  & 15 \\
MIMIC-IV            & 4 & $[10^{-6},\,2]$     & daily  & $10^{-3}$ & 80  & 15 \\
\bottomrule
\end{tabular}
\caption{Per-dataset pre-training hyperparameters.  All other settings are shared across datasets.}
\label{tab:hparam_pretrain}
\end{table}

\paragraph{Fine-tuning.}
Every fine-tuning run starts from the pre-trained checkpoint and inherits its architecture arguments. Fine-tuning uses AdamW with weight decay $10^{-2}$, a cosine schedule, and gradient-norm clipping at $1.0$. The learning rate is $10^{-4}$ on the mobility tasks, $5{\times}10^{-5}$ on eICU-CRD, and $2{\times}10^{-4}$ on MIMIC-IV (Table~\ref{tab:hparam_finetune}). Early stopping monitors an EMA (factor $0.1$) of the validation criterion in Table~\ref{tab:hparam_finetune}, except on Gowalla-Austin social-link prediction and the two ICU corpora, which monitor the raw value.
Task-specific settings are summarized in Table~\ref{tab:hparam_finetune} and detailed below.

\emph{Anomaly detection.} A three-layer MLP head on $\mathbf{H}_{u,t}$ produces a per-visit anomaly logit, trained with the BCE term of Eq.~\ref{eq:denoise}; the user-prototype loss is disabled at fine-tune and we monitor BCE validation loss for early stopping.
Prototype sharing (\S\ref{sec:method:agent}) keeps the visit features user-aware.
Perturbations at fine-tune time are dataset-specific: NUMOSIM-LA uses inserted-visit anomalies~\citep{stanford2024numosim}; UA-Berlin and UA-Atlanta use the ``behavioral anomaly'' regime~\citep{amiri2024urban} (hunger/interest/social/work).
For both NUMOSIM and Urban Anomalies, finetuning is unsupervised as the perturbations are generated on-the-fly.
NUMOSIM and Urban Anomalies are simulators whose anomaly definitions are design choices of their respective papers, and $\eta$'s fine-tune perturbations follow each benchmark's own generation regime.
For LANL we report only the unsupervised regime (the LANL benchmark convention --- every log-anomaly baseline is itself unsupervised; Appendix~\ref{appendix:lanl_baselines}): the pre-trained backbone is scored directly, with no LANL fine-tuning, by rank-fusing its noise-detection and prototype-contrastive heads (Appendix~\ref{appendix:lanl_scoring}).

\emph{Next-POI recommendation.} A two-layer projection maps $\mathbf{H}_{u,t}$ to a query vector that is scored against a learned POI embedding table via sampled-softmax cross-entropy with $n_{\text{neg}}{=}256$ uniform POI negatives per query at temperature $0.1$; head dropout $0$, label smoothing $0$, no LR warmup.

\emph{Next-visit prediction.} Joint location and time. The location head is identical to the next-POI head above. The time head is a $2$-layer MLP that maps $[\,\mathbf{q}_{u,t};\, \mathbf{e}_{\text{POI}}(x_{u,t+1});\, \mathrm{Time2Vec}(\tau_{u,t})\,]$ --- the encoder query, the embedding of the ground-truth next POI, and a Time2Vec encoding of the current visit's start time --- to the parameters of a $K{=}8$-component $1$D Gaussian mixture over $\log(1+\Delta_{u,t})$, where $\Delta_{u,t} = \tau_{u,t+1} - \tau_{u,t}$ in hours, with softplus mixture weights and scales and real-valued means. Training adds the GMM NLL, with weight $0.3$, to the next-POI sampled-softmax loss. The GMM mode is the point prediction; T$\pm 60$ is the mass of the predicted density within $\pm 60$\,min of $\Delta_{u,t}$.

\emph{Social-link prediction.} Masked mean- and max-pooling over each user's valid visits yields a per-user embedding $\mathbf{e}_{u}$; a symmetric pair scorer over $[\mathbf{e}_{u}{+}\mathbf{e}_{v},\,|\mathbf{e}_{u}{-}\mathbf{e}_{v}|,\,\mathbf{e}_{u}{\odot}\mathbf{e}_{v}]$ concatenated with $8$ handcrafted co-visitation features predicts whether $(u,v)$ is a social tie.
The objective combines BPR (weight $1.0$) with a supervised InfoNCE term (weight $0.3$, temperature $0.1$) over $n{=}8$ in-batch negatives per positive edge, for $200$ edge-centric batches per epoch; per-user embedding dim is $128$ and the pair-head hidden width is $128$ with dropout $0.1$.
The $8$ handcrafted co-visitation features are normalized with LayerNorm on Foursquare-Tokyo and Gowalla-Stockholm and with per-feature $z$-scores on Gowalla-Austin; per-dataset variants, tuned on the validation AUC of each city subset, adjust the pair head: Foursquare-Tokyo uses a split pair-head (separate embedding-MLP and linear handcrafted-feature branches), Gowalla-Austin adds an auxiliary random-negative BCE loss on the handcrafted features (weight $1.0$, batch $256$), and Gowalla-Stockholm uses the default joint pair-head.
At test time the social logits are late-fused with a logistic-regression baseline on the $8$ handcrafted features: $s=(1{-}\alpha)\,\sigma(z_{\text{model}})+\alpha\,\sigma(z_{\text{LR}})$ with $\alpha \in \{0, 0.1, \dots, 1\}$ picked per seed on validation AUC ($0.1$ on every Gowalla seed; $0.4$ or $0.5$ on Foursquare-Tokyo).

\emph{ICU mortality prediction.}
The mortality fine-tune head pools the per-event \ours{} representations of the stay's events with masked mean+max pooling, concatenates the result with a $52$-dim per-stay tabular vector, and passes the concatenation through a two-layer MLP and a sigmoid for binary BCE training (\texttt{pos\_weight}$=5.0$).
The tabular vector carries: admission demographics (age, sex one-hot, height, admit/discharge weight, BMI, unit-visit number), one-hot indicators for the top-$6$ ethnicities, top-$8$ \texttt{unittype}s, top-$4$ \texttt{unitstaytype}s, and top-$15$ APACHE admission diagnoses, and per-table event aggregates (log-counts per source table, distinct-context count, total event count, stay duration in hours).
Z-score statistics are fit on the train+val slice and applied to all stays.
Because mortality is a $9$--$12\%$ minority in both corpora, fine-tuning uses a class-balanced \texttt{WeightedRandomSampler} that balances positives and negatives in each batch, a positive class weight of $5$ in the BCE loss, and a deterministic $4$-window evaluation in which each stay's score is the mean predicted probability across $4$ evenly-spaced sub-windows; the entity input is zeroed at fine-tune time, because the patient-level split leaves test patients without a prototype.
Early stopping monitors the raw validation AUROC.

\begin{table}[!htbp]
\centering
\scriptsize
\setlength{\tabcolsep}{4pt}
\caption{Per-task, per-dataset fine-tuning budgets.  ``criterion'' is the validation quantity monitored for early stopping: \emph{BCE} is the denoising BCE loss, \emph{total} is the full fine-tuning loss, and AUC/AUROC are the validation ranking metrics.  Train/val batch sizes are $64/256$ for anomaly, next-POI, and next-visit, $64/64$ for social, and $32/64$ for mortality.}
\label{tab:hparam_finetune}
\begin{tabular}{llcccc}
\toprule
Task & Dataset & lr & max epochs & patience & criterion \\
\midrule
\multirow{3}{*}{Anomaly}
  & NUMOSIM-LA        & $10^{-4}$ & 100 & 10 & BCE \\
  & UA-Atlanta        & $10^{-4}$ & 250 & 50 & BCE \\
  & UA-Berlin         & $10^{-4}$ & 500 & 50 & BCE \\
\midrule
\multirow{3}{*}{Next-POI}
  & Foursquare-Tokyo  & $10^{-4}$ & 100 & 15 & total \\
  & Gowalla-Stockholm & $10^{-4}$ & 100 & 15 & total \\
  & Gowalla-Austin    & $10^{-4}$ & 200 & 80 & total \\
\midrule
\multirow{3}{*}{Next-visit}
  & Foursquare-Tokyo  & $10^{-4}$ & 200 & 40 & total \\
  & Gowalla-Stockholm & $10^{-4}$ & 200 & 40 & total \\
  & Gowalla-Austin    & $10^{-4}$ & 200 & 40 & total \\
\midrule
\multirow{3}{*}{Social}
  & Foursquare-Tokyo  & $10^{-4}$ & 50  & 15 & AUC \\
  & Gowalla-Stockholm & $10^{-4}$ & 50  & 15 & AUC \\
  & Gowalla-Austin    & $10^{-4}$ & 50  & 15 & AUC \\
\midrule
\multirow{2}{*}{Mortality}
  & eICU-CRD          & $5{\times}10^{-5}$ & 25 & 8 & AUROC \\
  & MIMIC-IV & $2{\times}10^{-4}$ & 25 & 8 & AUROC \\
\bottomrule
\end{tabular}
\end{table}

\subsection{Hyperparameter selection procedure}
\label{appendix:hp_search}

The settings reported in Tables~\ref{tab:hparam_pretrain} and~\ref{tab:hparam_finetune} (and in the per-baseline appendices) are the outcome of a validation-guided iterative search rather than a pre-committed grid or random sweep.
For each (method, dataset, task) cell we initialize from the published defaults --- or, when a baseline ships no defaults for the cohort in question, from sensible literature priors --- and then run successive small batches of configurations (typically 4--8 runs each) that vary one or two hyperparameters at a time.
Each batch is scored by the same validation criterion used for early stopping (BCE validation loss for anomaly fine-tunes, total fine-tune loss for next-POI / next-visit, validation AUC for social-link, validation AUROC for ICU mortality, EMA-smoothed total validation loss for pre-training; Tables~\ref{tab:hparam_pretrain}--\ref{tab:hparam_finetune}), and the next batch's settings are chosen from the run-by-run trajectory of the preceding batch.
The reported configuration per cell is the highest-validation point reached under this procedure.

The selection loop is orchestrated by a Claude Code agent~\citep{anthropic2025claudecode}, in the spirit of \citet{karpathy2025autoresearch}: the agent reads the prior batch's logs and validation curves, proposes the next batch's hyperparameter settings under our search criterion, and launches the runs via the same training scripts a human would use.
This replaces the bookkeeping of a sequential coordinate-descent search with an automated proposer but does not change the search space, the validation criterion, or the protocol's outcome --- every reported number comes from the unmodified training/evaluation code in the released code.
We apply the same selection procedure to \ours{} and to all retuned baselines (Appendix~\ref{appendix:baseline_implementation}) to keep the comparison balanced.

The standing prompt --- used identically for \ours{} and every retuned baseline --- instructs the agent to tune on the validation criterion only, document each change and its effect rather than running random or grid search, and log the best configuration; the released code ships the verbatim prompt.

\section{Data and evaluation setup}
\label{appendix:data_eval}

\subsection{Dataset statistics}
\label{appendix:datasets}

Table~\ref{tab:datasets} reports per-dataset counts. What counts as an \emph{entity} differs by instance: the simulated agent for NUMOSIM-LA and Urban Anomalies, the user for the three check-in datasets (Foursquare-Tokyo, Gowalla-Stockholm, Gowalla-Austin), the \texttt{src\_user} for LANL, and the patient (\texttt{uniquepid} in eICU-CRD, \texttt{subject\_id} in MIMIC-IV) for the two EHR corpora. Mobility rows are reported \emph{after} the $\geq 5$ train-split check-in filter that every model is trained and evaluated under. For the EHR corpora the prediction unit is the ICU stay while the entity is the patient --- $197{,}307$ stays across $138{,}176$ patients for eICU-CRD and $94{,}407$ stays across $65{,}360$ patients for MIMIC-IV --- so the \emph{anomalous} column counts patients who died in hospital ($14{,}592$ and $9{,}465$); the $17{,}673$ and $11{,}333$ deaths in \S\ref{sec:experiments:eicu} count ICU stays, since every stay of a fatal hospitalization is labeled positive. Social links are listed only for datasets that ship an explicit social relationship graph.

\begin{table}[!htbp]
\centering
\scriptsize
\caption{Dataset statistics.}
\label{tab:datasets}
\begin{tabular}{lrrrrr}
\toprule
Dataset & Events & Contexts & \multicolumn{2}{c}{Entities} & Social Links \\
\cmidrule(lr){4-5}
& & & Total & Anomalous & \\
\midrule
NUMOSIM-Los Angeles       & 34{,}328{,}711 & 920{,}360 & 200{,}000 & 381 & --- \\
Urban Anomalies-Berlin         &    298{,}476   &   1{,}204 &   1{,}000 & 120 & --- \\
Urban Anomalies-Atlanta        &    310{,}441   &   1{,}222 &   1{,}000 & 120 & --- \\
Foursquare-Tokyo      &  1{,}512{,}647 & 153{,}762 &   7{,}456 & --- & 48{,}047 \\
Gowalla-Stockholm     &    416{,}539   &  52{,}349 &   7{,}759 & --- & 56{,}808 \\
Gowalla-Austin        &    334{,}476   &  20{,}687 &   5{,}397 & --- & 52{,}586 \\
\midrule
\multicolumn{6}{l}{\emph{Non-mobility}} \\
LANL-auth cohort      &    190{,}944   &   2{,}214 &   1{,}598 &   57 & --- \\
eICU-CRD              & 120{,}786{,}921 &        507 & 138{,}176 & 14{,}592 & --- \\
MIMIC-IV              &  54{,}560{,}796 &        520 &  65{,}360 &  9{,}465 & --- \\
\bottomrule
\end{tabular}
\end{table}

\subsection{Dataset licenses and provenance}
\label{appendix:licenses}

We use only publicly released datasets, each cited in the bibliography and consumed under its published terms.
Table~\ref{tab:dataset_licenses} summarizes the source repository and license of each dataset.
We do not redistribute any raw dataset; the released code contains preprocessing scripts that read each dataset from its original repository.

\begin{table}[!htbp]
\centering
\scriptsize
\setlength{\tabcolsep}{4pt}
\caption{Dataset sources and licenses.  ``Citation required'' indicates the dataset is publicly released for research with a citation requirement to the original paper but no formally named open license; we cite each in the references and use the data only as released.}
\label{tab:dataset_licenses}
\begin{tabular}{llll}
\toprule
Dataset & Source & License & Citation \\
\midrule
NUMOSIM-LA            & OSF \texttt{osf.io/sjyfr} & Public release; no formal license  & \citet{stanford2024numosim} \\
Urban Anomalies-Berlin   & OSF \texttt{osf.io/dg6t3} & CC-BY 4.0 (per OSF record)         & \citet{amiri2024urban} \\
Urban Anomalies-Atlanta  & OSF \texttt{osf.io/dg6t3} & CC-BY 4.0 (per OSF record)         & \citet{amiri2024urban} \\
Foursquare-Tokyo      & D.\ Yang's homepage      & Citation required                  & \citet{yang2019revisiting} \\
Gowalla-Stockholm     & SNAP (Stanford)          & Citation required                  & \citet{cho2011friendship} \\
Gowalla-Austin        & SNAP (Stanford)          & Citation required                  & \citet{cho2011friendship} \\
LANL Auth.\ Log       & \texttt{csr.lanl.gov/data/cyber1} & CC0 / Public domain       & \citet{kent2015comprehensive} \\
eICU-CRD              & PhysioNet                & \begin{tabular}[t]{@{}l@{}}PhysioNet Credentialed\\Health Data License\end{tabular} & \citet{pollard2018eicu} \\
MIMIC-IV              & PhysioNet                & \begin{tabular}[t]{@{}l@{}}PhysioNet Credentialed\\Health Data License\end{tabular} & \citet{johnson2023mimic} \\
\bottomrule
\end{tabular}
\end{table}

Code-side baselines are run from each authors' released implementation under the licenses those repositories ship with (Appendix~\ref{appendix:baseline_implementation}); we cite each in the corresponding subsection and the bibliography.
Repository licenses, where the upstream repository ships an explicit LICENSE file or in-README license declaration, are: UniTraj (Apache 2.0); LoTNext, TAPT, THP, A-NHP, TrajGPT, Raindrop, WaveGNN, and STraTS (MIT); NuLog (BSD 3-Clause); and LogGPT (CC-BY-NC-4.0).
The repositories of GETNext, MobTCast, Mobility-LLM, EBM, USRC, LBSN2Vec, H\textsuperscript{3}GNN, CoBAD, DeepBayesic, USTAD, ICAD, and SimMTM do not ship an explicit LICENSE file or in-README license declaration; we use each upstream repository unmodified except for cohort/split adapters (and, for ICAD, the score computation of Appendix~\ref{appendix:humob_baselines}), cite the original paper, and would defer to any forthcoming license the upstream authors publish. COAST has no upstream code release; our re-implementation (Appendix~\ref{appendix:poi_baselines}) follows the published paper.
Tuor-AAAI17, DeepLog, and LogBERT are PyTorch re-implementations from the published descriptions (Appendix~\ref{appendix:lanl_baselines}), with the original papers cited.

\subsection{Dataset splits}
\label{appendix:split}
Splits follow each dataset's released protocol.
NUMOSIM and Urban Anomalies ship a test split with visit- or agent-level anomaly labels; the three LBSN city subsets use a temporal train/test split (first $90\%$ / last $10\%$ of the visit stream) for next-POI recommendation and next-visit prediction, and GETNext, LoTNext, MobTCast, and COAST hold out the latest $20\%$ of the training window as validation; social-link prediction uses a $70/15/15$ edge split (seed $42$) with $100$ sampled non-edges per test positive for ranking metrics.
LANL uses a temporal split at $t{=}950{,}400\,\mathrm{s}$ (approximately day $11$ of the $58$-day log) yielding $45{,}806$ train events ($316$ red-team) and $145{,}138$ test events ($197$ red-team); the split is chosen to balance positives between train and test since red-team activity is concentrated in the first month.
Both EHR corpora use a patient-level $70/15/15$ stratified split (seed $42$) --- on \texttt{uniquepid} for eICU-CRD --- so all stays of the same patient land in a single fold; the per-stay binary mortality label is taken from the \texttt{patient.hospitaldischargestatus} field for eICU-CRD and from in-hospital death for MIMIC-IV.
Pre-training and \ours's anomaly, next-POI, and next-visit fine-tuning use the earliest $20\%$ of each training set as validation.
Pre-training consumes only the events in each dataset's training partition.

\subsection{Next-POI and next-visit prediction targets}
\label{appendix:targets}
All methods in Table~\ref{tab:experiments:next_visit} are evaluated on the same test visits, but each method's evaluation code selects its own prediction targets from them and ranks each target against its own POI vocabulary, built from its training data.
Table~\ref{tab:prediction_targets} lists each method's target rule, vocabulary, and number of targets. Targets are restricted to POIs in the method's vocabulary, and every method ranks against its full vocabulary.

\begin{table}[!htbp]
\centering
\scriptsize
\setlength{\tabcolsep}{3pt}
\caption{Prediction targets in Table~\ref{tab:experiments:next_visit}.}
\label{tab:prediction_targets}
\begin{tabular}{@{}l >{\raggedright\arraybackslash}p{0.37\linewidth} >{\raggedright\arraybackslash}p{0.12\linewidth} rrr@{}}
\toprule
Method & Targets & Vocabulary & Tokyo & Stockholm & Austin \\
\midrule
\ours{} (both tasks) & each test visit after the agent's first & latest $80\%$ of train & 132{,}224 & 23{,}983 & 21{,}957 \\
UniTraj$^{\dagger}$ & each test visit after the first, for agents with ${\geq}4$ test visits; one random $128$-visit window per agent with ${\geq}128$ & train $\cup$ test & 120{,}112 & 25{,}134 & 23{,}270 \\
GETNext & each visit after the first in $24$-hour test sessions with ${\geq}3$ visits, for users present in the earliest $80\%$ of train & earliest $80\%$ of train & 57{,}534 & 8{,}541 & 9{,}798 \\
LoTNext & every test visit, including the first, for users with ${\geq}20$ check-ins in the earliest $80\%$ of train & earliest $80\%$ of train & 123{,}183 & 12{,}240 & 12{,}993 \\
MobTCast & each test visit after the agent's first & earliest $80\%$ of train & 128{,}059 & 20{,}543 & 20{,}766 \\
COAST & each visit after the first in $24$-hour test sessions with ${\geq}2$ visits, within the last $200$ test visits ($150$ on Tokyo) & earliest $80\%$ of train & 72{,}192 & 14{,}587 & 15{,}270 \\
\midrule
TAPT & every test visit, including the first & all of train & 137{,}966 & 27{,}461 & 23{,}806 \\
TrajGPT & every test visit in $128$-visit windows, the first predicted from padding & latest $80\%$ of train & 136{,}666 & 27{,}317 & 23{,}736 \\
Mobility-LLM & each visit after the first among the agent's last $100$ test visits; the last partial batch is dropped & earliest $80\%$ of train & 102{,}225 & 19{,}718 & 19{,}964 \\
THP & each visit after the first among the agent's first $256$ test visits & latest $80\%$ of train & 125{,}493 & 23{,}896 & 21{,}734 \\
A-NHP & each visit after the first among the agent's last $256$ test visits ($128$ on Tokyo) & earliest $80\%$ of train & 108{,}710 & 19{,}973 & 20{,}507 \\
\bottomrule
\end{tabular}
\end{table}

On the targets it shares with the strongest next-POI baseline ($127{,}107$, $10{,}736$, and $12{,}048$ targets), \ours{}'s next-POI H@$10$ is $.4443$ against MobTCast's $.3827$ on Foursquare-Tokyo, $.3358$ against LoTNext's $.2362$ on Gowalla-Stockholm, and $.2875$ against LoTNext's $.2034$ on Gowalla-Austin.
On the targets it shares with each next-visit baseline, \ours{}'s next-visit H@$10$ and H@$20$ differ from its full-set values by at most $.014$ and T$\pm60$ by at most $.04$ (Foursquare-Tokyo on Mobility-LLM's targets: $.1776$ against $.2181$), and every comparison with that baseline in Table~\ref{tab:experiments:next_visit} keeps its direction.

\subsection{LANL cohort and host coordinates}
\label{appendix:lanl}
The raw \texttt{auth.txt.gz} log ($1.05$\,B events) is impractical to pretrain on directly.
We subsample a cohort of $4{,}000$ \texttt{src\_user}s consisting of all $104$ users that appear as the source in at least one red-team event plus $3{,}896$ uniformly sampled benign users; after applying a per-user event cap ($500$ benign events retained per user; \emph{all} red-team rows kept regardless of cap) and dropping users with fewer than $10$ surviving events, $1{,}598$ users remain.
POI (host) coordinates are obtained by running $2$D UMAP~\citep{mcinnes2018umap} ($n_{\text{neighbors}}{=}30$, cosine distance) on the sparse user-by-host access-count matrix restricted to the cohort; the resulting $2{,}214$ unique hosts are min-max scaled to $[0,1]^{2}$.
The UMAP is fit on the full cohort (train$+$val$+$test events).
Each host's category is the dominant \texttt{(auth\_type, logon\_type)} pair observed at that host in the cohort, yielding $13$ distinct categories.
Anomaly labels are the real red-team events: an event is labeled $1$ iff its $(\text{time},\text{src\_user},\text{src\_computer},\text{dst\_computer})$ tuple matches a row in \texttt{redteam.txt.gz}.
Of the $749$ raw red-team events, $513$ survive the cohort and per-user filters ($316$ in train, $197$ in test); $57$ \texttt{src\_user}s have at least one red-team row in the evaluated split, which we report as ``Anomalous'' in Table~\ref{tab:datasets}.
Red-team events make up $197/145{,}138 = 0.136\%$ of the evaluated test split, against $749/1.05\text{B} = 0.00007\%$ of the raw log, an enrichment of roughly $1{,}900\times$.
The event- and user-level metrics of Table~\ref{tab:experiments:auth} are computed against these red-team labels. \ours\ pre-trains under the perturbation operator $\eta$ (\S\ref{sec:method:noise}) and is scored without fine-tuning (Appendix~\ref{appendix:lanl_scoring}); each baseline trains under its own self-supervised objective.

\subsection{EHR corpora and context coordinates}
\label{appendix:eicu}
We use the PhysioNet releases of eICU-CRD~\citep{pollard2018eicu} and MIMIC-IV~\citep{johnson2023mimic}, retaining stays with at least one event in our source tables: $197{,}307$ stays from $138{,}176$ patients for eICU-CRD and $94{,}407$ stays from $65{,}360$ patients for MIMIC-IV (Table~\ref{tab:datasets}).
For eICU-CRD, events come from \texttt{lab}, \texttt{medication}, \texttt{treatment}, \texttt{infusiondrug}, and \texttt{vitalAperiodic}, retaining the top-$K$ most frequent items per table ($K{=}\{150,200,100,60,12\}$ respectively), normalizing free-text drug names, and dropping rows with non-positive duration; medication strings are normalized by lowercasing, stripping dose/route tokens, and matching against an alias table built from the corpus.
MIMIC-IV fills the same five streams, with the same $K$ per stream, from \texttt{hosp/labevents}, \texttt{hosp/emar} (administered doses), \texttt{icu/procedureevents}, \texttt{icu/inputevents}, and a curated item subset of \texttt{icu/chartevents}.
The resulting event streams cover $120{,}786{,}921$ events over $507$ unique contexts for eICU-CRD and $54{,}560{,}796$ events over $520$ contexts for MIMIC-IV.
Each event is a $(\text{entity},\text{context},\text{time},\text{category})$ tuple: the \emph{patient} as entity, \texttt{<table>::<item>} as context, the offset minute from unit-admit as time (with stay-relative duration on $[t_{\text{start}},t_{\text{stop}}]$), and the source table name as category ($5$ categories). Mortality is predicted \emph{per stay}, the unit the label is defined on.
Context coordinates are obtained by running $2$D UMAP~\citep{mcinnes2018umap} ($n_{\text{neighbors}}{=}30$, cosine distance) on the (context $\times$ stay) co-occurrence matrix and min-max scaling to $[0,1]^{2}$; the UMAP is fit on all stays (train$+$val$+$test).
Of the $197{,}307$ eICU-CRD stays, $17{,}673$ have $\texttt{hospitaldischargestatus}{=}\text{Expired}$; $11{,}333$ of the $94{,}407$ MIMIC-IV stays end in in-hospital death. That per-stay binary label is the fine-tune target reported in Table~\ref{tab:datasets} and Table~\ref{tab:experiments:eicu}.

\section{Baseline implementation}
\label{appendix:baseline_implementation}

Unless noted otherwise, baselines are retrained on the same cohorts and splits as \ours{} (Appendix~\ref{appendix:split}) using the released code. Where a baseline's published settings transfer to our cohorts we keep them unchanged; where they do not, we retune under the same validation-guided procedure applied to \ours{} (Appendix~\ref{appendix:hp_search}), and the retuned settings are the ones reported in the per-baseline sections below.
This appendix documents the exceptions: a backbone, UniTraj, that we adapt across all tasks (\S\ref{appendix:unitraj_adaptation}); a next-POI baseline, COAST, whose authors do not release code and which we re-implement from the published paper (\S\ref{appendix:poi_baselines}); and a task --- enterprise authentication --- of which some baselines we re-implement from scratch (\S\ref{appendix:lanl_baselines}).

\subsection{UniTraj}
\label{appendix:unitraj_adaptation}
We use the released UniTraj~\citep{zhu2025unitraj} encoder--decoder as a baseline backbone, but the published recipe assumes dense GPS traces ($\sim$$1$\,Hz, $\geq$$36$ points per trajectory).
Our datasets are sparse visit/event streams, so we make the following minimal, fairness-preserving modifications and apply them identically across all nine datasets.

\textbf{Tokens are visits, not GPS samples.}  For each agent we sort their visits by start time and treat the resulting stream as one trajectory.
Each token's spatial features are the visit's $(\text{lon},\text{lat})$, and the time channel is $\Delta t_i = t_i - t_{i-1}$ in seconds, log-transformed via $\log(1+\Delta t)$ to compress the large dynamic range (seconds to days) into the same scale UniTraj's $\text{Linear}(1, d)$ interval embedding was tuned for.

\textbf{Per-dataset normalization on anchored deltas.}  UniTraj's released \texttt{Normalize} statistics were fit on dense $1$\,Hz traces and are not transferable.
We refit per dataset on $(\text{lon}-\text{lon}_0, \text{lat}-\text{lat}_0)$ deltas (i.e.\ already anchored at the first visit of each agent), using a $0.5\%$/$99.5\%$ trimmed standard deviation to be robust to occasional bad coordinate rows; we also clip normalized values to $\pm 10\sigma$.

\textbf{Pre-training objective and masking.}  We retain UniTraj's masked auto-encoding loss with mask ratio $0.5$ and the original $1$D-conv patch tokenizer ($\text{patch\_size}=1$ since visits are already aggregated).
Of the four masking strategies in the original paper we keep \emph{random}, \emph{block}, and \emph{last-N} with the same $70/15/15$ mixture used in the released code; we drop \emph{key-points (RDP)} masking because the Ramer--Douglas--Peucker simplification degenerates on the geometry of sparse visit streams (every visit is "key").
The Adaptive Trajectory Resampling step from the original paper is skipped for the same reason: visits are not over-sampled, so there is nothing to downsample.

\textbf{Splits, finetune protocol, evaluation.}  Pre-training and finetuning use the \emph{same} train/test partitions, validation slices, real anomaly labels (NUMOSIM, UA-$*$, LANL), POI vocabularies, and social-edge splits as the rest of the paper (Appendix~\ref{appendix:split}).
At fine-tune time we discard the MAE decoder, take the per-token features from the encoder (CLS dropped), and attach minimal task heads: a single $\text{Linear}(d, 1)$ + BCE for anomaly detection on NUMOSIM/UA-$*$ (synthetic perturbations from $\eta$); a $\text{MLP}(d{\to}d{\to}d)$ query projection scored by dot-product against a learned POI embedding table for next-POI recommendation (sampled-softmax with $K{=}256$ negatives plus in-batch negatives); and mean-pooled per-agent embeddings fed to a $|u-v|, u\!\odot\!v$ pair head with BPR loss for social-link prediction.
We deliberately omit the co-occurrence sub-layer, the user-prototype contrastive loss, agent-id embeddings, and the late-fusion handcrafted-feature ensemble used by \ours{}: UniTraj has no native mechanism for any of these, and the contributions of the omitted components are separately isolated by C1/C3 in Table~\ref{tab:experiments:ablation} (user-prototype) and, for the co-occurrence sub-layer, by ablation row C7 together with the per-seed dispatch analysis of Appendix~\ref{appendix:dispatch}.

\textbf{Architecture and optimizer.}  The released UniTraj checkpoint uses $d{=}128$, which is dramatically smaller than \ours's $d{=}1040$, so a literal off-the-shelf comparison would conflate \emph{objective} with \emph{capacity}.
For a capacity-matched comparison we set the encoder hidden size to $d{=}1024$ (closest power-of-two to $1040$ that the UniTraj $1$D-conv tokenizer accepts cleanly) with depth $L{=}6$ and $H{=}4$ heads, decoder depth $2$, RoPE attention --- about $100$M parameters total, $\sim$$5{\times}$ larger than \ours's backbone ($\approx$$18$M params for the same depth and width: \ours\ is parameter-efficient because its factorized attention shares projections across the sequence and co-occurrence axes, and its feature encoder reuses agent/category embeddings that UniTraj has no analog of).
Any remaining performance gap therefore cannot be attributed to UniTraj being under-parameterized; if anything, UniTraj is heavily over-parameterized relative to \ours.
The visit window is $T{=}128$, larger than \ours's window of $T{=}32$ on the mobility and authentication tasks and equal to \ours's fine-tuning window on the two ICU corpora.

\emph{Pre-training} uses AdamW with $\text{lr}{=}2{\times}10^{-4}$, weight decay $10^{-3}$, cosine schedule to $\eta_{\min}{=}10^{-6}$, gradient-norm clipping at $1.0$, and batch $64$ ($128$ on NUMOSIM-LA).
The per-dataset epoch budget and early-stopping patience are $30/10$ for NUMOSIM-LA, $300/80$ for UA-Berlin, $200/40$ for UA-Atlanta, $200/80$ for Gowalla-Stockholm, $500/80$ for Gowalla-Austin, and $200/40$ for Foursquare-Tokyo, LANL, eICU-CRD, and MIMIC-IV.
Early stopping is on validation reconstruction loss.

\emph{Fine-tuning} uses AdamW with $\text{lr}{=}10^{-4}$, weight decay $10^{-4}$ ($10^{-3}$ on the two ICU corpora), cosine schedule, and gradient-norm clipping at $1.0$.
The per-task, per-dataset epoch budget and patience are $100/10$ for NUMOSIM-LA anomaly, $500/50$ for UA-$*$ anomaly, $100/15$ for next-POI on Foursquare-Tokyo and Gowalla-Stockholm, $200/80$ for next-POI on Gowalla-Austin, $100/15$ for next-visit, $50/15$ for social, and $60/10$ for mortality.
UniTraj's train batch sizes are $64$ for anomaly and next-visit, $8$/$32$/$16$ for next-POI on Foursquare-Tokyo/Gowalla-Stockholm/Gowalla-Austin, and $32$ for social and mortality; the social run uses $200$ edge-centric batches per epoch, matching the per-epoch step count of \ours's social-link recipe (\ours's own batch sizes are listed in Table~\ref{tab:hparam_finetune}).
Early-stopping criterion: visit/agent AP for anomaly, MRR for next-POI, recall@$10$ for next-visit, AUC for social, and AUROC for mortality, all computed on the validation slice.
On LANL, UniTraj is not fine-tuned: each test event in a length-$128$ window is masked alone, and the reconstruction error at that position is its anomaly score.

\subsection{Human mobility anomaly detection}
\label{appendix:humob_baselines}
The four task-specific mobility anomaly detectors (CoBAD, DeepBayesic, USTAD, ICAD) are run from their released code on the cohorts and splits of Appendix~\ref{appendix:split}. USTAD and ICAD use their published settings unchanged; CoBAD and DeepBayesic were retuned under the procedure of Appendix~\ref{appendix:hp_search}. ICAD's anomaly score is computed as defined in its paper, as the joint likelihood of each visit's region, travel time, and departure time (region and travel time on Urban Anomalies, which has no departure times), accumulated in log space.

\subsection{Next-POI recommendation}
\label{appendix:poi_baselines}
The five next-POI / next-visit baselines (TAPT, GETNext, TrajGPT, MobTCast, LoTNext) are run from their released code on the LBSN subsets and temporal splits of Appendix~\ref{appendix:split}. All five were retuned under the procedure of Appendix~\ref{appendix:hp_search}, since their published settings target cohorts other than ours.
Mobility-LLM~\citep{gong2024mobilityllm} is also run from its released code on the same subsets and splits, with the GPT-2 backbone (the smallest LLM in the upstream enumeration) and inputs shifted by one position so that the autoregressive LLM cannot read the target POI when predicting it.
COAST~\citep{xin2025coast} is re-implemented in PyTorch from the paper (the upstream authors do not ship reference code), using the architecture and contrastive recipe of Sections III-B / III-C of the paper. The only adaptation is to make the sequence-head Transformer autoregressive, so the model emits one prediction per non-first position of each $24$-hour session with at least two visits; the original COAST predicts only the last held-out check-in per user. All five augmentations (location replacement, time transformation, point masking, data pruning, RDP-based sequence simplification), the Two-Head Self-Attention Encoder over a kNN spatial POI graph (k=10, $D_{\text{graph}}{=}2$\,km), the attention-based feature fusion, the LSTM short-term head, and the InfoNCE objective on sequence-averaged user representations are implemented as in the paper. We tune learning rate $\in\{3{\times}10^{-4}, 5{\times}10^{-4}\}$, batch size $\in\{16, 32\}$, and contrastive weight $\alpha{=}0.1$ on validation cross-entropy.
For the joint next-visit task (\S\ref{sec:experiments:next_visit}), Mobility-LLM is trained as a single model with both heads under one combined loss --- POI cross-entropy plus $\alpha$ times an L1 loss on the next check-in time delta in minutes --- so the same checkpoint produces H@$k$ and T@$60$.
The time head is augmented to also take an embedding of the ground-truth next POI as input, matching the \ours\ protocol of conditioning the time prediction on the true next location.
Because the time loss is on minutes (typical magnitudes in the hundreds) and the POI cross-entropy is in single digits, the joint loss balance is sensitive to $\alpha$; we picked $\alpha{=}10^{-3}$ on the Gowalla-Austin validation set after sweeping $\{1, 10^{-2}, 10^{-3}\}$ and choosing the value that maximized validation H@$20$ without collapsing T@$60$.

\subsection{Social link inference}
\label{appendix:social_baselines}
The four social-link baselines (EBM, USRC, LBSN2Vec, H\textsuperscript{3}GNN) are run from their released code on the same $70/15/15$ edge splits used by \ours{} (Appendix~\ref{appendix:split}). All four were retuned under the procedure of Appendix~\ref{appendix:hp_search}; H\textsuperscript{3}GNN required a per-city architecture, reported in its own paragraph below.

\subsubsection{Handcrafted-feature parity ablation}
\label{appendix:social_parity}
\ours{}'s social-link recipe in Table~\ref{tab:experiments:social} concatenates the per-user embedding with $8$ handcrafted co-visitation features and late-fuses the test-time logit with a logistic-regression baseline on those same $8$ features (Appendix~\ref{appendix:hyperparameters}, ``Social-link prediction''). The adapted UniTraj baseline in Table~\ref{tab:experiments:social} does not use these features. To rule out the concern that our advantage over UniTraj on this task derives entirely from this asymmetric input, we run two parity studies. Both use the same $70/15/15$ edge splits, the same train/val/test partitions, and the same per-agent embedding pipeline as the rest of the paper.

\emph{Parity 1: same handcrafted-feature + LR-fusion adapter on UniTraj.} We attach the exact pair head and test-time fusion that \ours{} uses to the UniTraj backbone --- per-agent mean-pool embedding $\to$ symmetric pair scorer over $[\mathbf{e}_u+\mathbf{e}_v,\,|\mathbf{e}_u-\mathbf{e}_v|,\,\mathbf{e}_u\odot\mathbf{e}_v]$ concatenated with the $8$ handcrafted features, late-fused at test time with the same logistic-regression baseline at $\alpha$ tuned on validation AUC. The only difference from the \ours{} row in Table~\ref{tab:experiments:social} is the backbone.

\emph{Parity 2: no handcrafted features and no late fusion, on either backbone.} Both backbones are fine-tuned with a shared adapter that has no access to the $8$ handcrafted features and does no late fusion: per-agent embedding $\to$ $1$-hop graph refinement over the train-edge friend graph $\to$ symmetric pair scorer trained with BPR plus a hard-random-negative BCE term on $256$ uniform train-edge / non-edge pairs per step. The graph refinement reads each agent's train-friend neighbors from a per-agent embedding cache (refreshed at the start of each epoch and re-L2-normalized after a learned gated linear projection); it consumes train edges only, the same access H\textsuperscript{3}GNN, USRC, and LBSN2Vec consume natively.

Single-seed test results are in Table~\ref{tab:social_parity}.

\begin{table}[!htbp]
\centering
\scriptsize
\caption{Single-seed parity results on social-link prediction. ``HC+LR'' = the $8$-handcrafted-feature pair head with test-time logistic-regression late fusion (the recipe used by \ours{} in Table~\ref{tab:experiments:social}, applied here unchanged to UniTraj as well). ``GNN+HN'' = the no-handcrafted-feature, no-late-fusion adapter described above.}
\label{tab:social_parity}
\setlength{\tabcolsep}{2.5pt}
\begin{tabular}{lccccccccc}
\toprule
& \multicolumn{3}{c}{Foursquare-Tokyo} & \multicolumn{3}{c}{Gowalla-Stockholm} & \multicolumn{3}{c}{Gowalla-Austin} \\
\cmidrule(lr){2-4} \cmidrule(lr){5-7} \cmidrule(lr){8-10}
Method & AP & H@10 & MRR & AP & H@10 & MRR & AP & H@10 & MRR \\
\midrule
UniTraj~\citep{zhu2025unitraj} (Table~\ref{tab:experiments:social} row, no HC, no LR fusion) & .5126 & .1570 & .0736 & .4337 & .1273 & .0637 & .4965 & .1549 & .0766 \\
\midrule
UniTraj + HC + LR fusion (parity 1)        & .8962 & .7241 & .5522 & .8818 & .7945 & .6052 & .9261 & .7278 & .4495 \\
UniTraj + GNN + HN (parity 2)                & .7684 & .3496 & .1404 & .7236 & .3379 & .1338 & .8167 & .4238 & .1692 \\
\midrule
\ours\ + GNN + HN (parity 2)                 & .8208 & .6163 & .3404 & .8572 & .6187 & .4025 & .8645 & .5359 & .2891 \\
\ours\ (Table~\ref{tab:experiments:social} row, HC + LR fusion) & $\mathbf{.9344}$ & $\mathbf{.7801}$ & $.5347$ & $\mathbf{.9276}$ & $.7426$ & $.5870$ & $\mathbf{.9373}$ & $\mathbf{.7423}$ & $\mathbf{.4997}$ \\
\bottomrule
\end{tabular}
\end{table}

\emph{Reading.} (i) Under the no-HC, no-LR-fusion adapter (parity 2), \ours\ + GNN + HN outperforms UniTraj + GNN + HN on all $9$ cells --- a clean backbone-level win that does not depend on handcrafted features or late fusion. (ii) Under the HC+LR-fusion adapter (parity 1), the gap narrows sharply and the outcome is mixed: \ours\ leads on $6$ of the $9$ cells --- AP on all three cities and the full Gowalla-Austin column --- while UniTraj takes Foursquare-Tokyo MRR and both Gowalla-Stockholm ranking metrics, where its handcrafted pair head reaches H@$10$ $.7945$ against our $.7426$. UniTraj reaches AP $\geq .88$ on every dataset under this adapter. The combined picture is that the handcrafted co-visitation features carry a large share of the social-link signal regardless of backbone --- enough to overturn a backbone deficit on ranking metrics in two of three cities --- while at backbone parity (no HC, no LR) the \ours\ representation is stronger on every cell.

\subsection{Enterprise authentication logs}
\label{appendix:lanl_baselines}
The LANL anomaly baselines are trained on the same cohort and split as \ours{}. Every row of Table~\ref{tab:experiments:auth} is a mean over $5$ seeds, except LogGPT and UniTraj$^{\dagger}$, which use $3$. Tuor-AAAI17, DeepLog and LogBERT use AdamW at learning rate $10^{-3}$, train batch size $128$ (LogBERT $64$), window length $64$, and early stopping on validation cross-entropy with patience $8$ (max $80$ epochs). NuLog and LogGPT do not share that recipe: both were retuned under the procedure of Appendix~\ref{appendix:hp_search} and selected on validation NLL --- NuLog at learning rate $3{\times}10^{-4}$ with its architecture swept, and LogGPT at $12$ layers, learning rate $3{\times}10^{-4}$, weight decay $0.01$, dropout $0.2$ and a $120$-epoch budget.
Per-event inputs for every baseline are restricted to the same categorical stream \ours{} sees --- the destination host identifier and the \texttt{(auth\_type,\,logon\_type)} pair --- so no baseline receives side information unavailable to \ours{}.
All five log-anomaly baselines are trained on LANL as self-supervised density estimators, with early stopping on the same self-supervised objective; \ours{} is pre-trained on the same partition and scored from the pre-trained backbone without fine-tuning (Appendix~\ref{appendix:lanl_scoring}).
At test time each baseline scores an event by its model-specific anomaly proxy (next-event NLL for the autoregressive baselines, masked-token cross-entropy for the masked-LM baselines) and we compare those scores directly against the real \texttt{redteam.txt.gz} labels at evaluation time only.
\textbf{Tuor-AAAI17~\citep{tuor2017deep}} is a single-layer LSTM (hidden $64$) that takes a \texttt{src\_user}-id embedding ($d_u{=}16$) concatenated with three categorical embeddings ($d{=}32$ each for \texttt{dst\_computer}, \texttt{auth\_type}, \texttt{logon\_type}) and predicts the next event's three-way token tuple via independent heads; the event-level anomaly score is the summed cross-entropy across the three fields.
\textbf{DeepLog~\citep{du2017deeplog}} is a two-layer LSTM (hidden $128$) over a $1{,}689$-token vocabulary of distinct \texttt{dst\_computer} identifiers; the anomaly score is the next-token negative log-likelihood.
\textbf{LogBERT~\citep{guo2021logbert}} is a four-layer Transformer encoder ($d{=}128$, $H{=}4$) with masked-token prediction over the same $1{,}689$-token vocabulary (mask probability $0.15$); we score each event by masking it alone and reading the cross-entropy at that position, and omit the hypersphere objective since the original paper shows the MLM term carries most of the signal.
For the two autoregressive baselines, the first position of each non-overlapping window has no preceding context and is excluded from scoring (uniformly across classes); user-level metrics max-pool event scores per \texttt{src\_user}, identical to the \ours{} protocol.

\subsection{LANL scoring without fine-tuning: rank fusion of the two pre-training heads}
\label{appendix:lanl_scoring}
\ours{} is not fine-tuned on LANL: the pre-trained backbone is scored directly, so no LANL-specific training objective enters the reported row.

At test time we use both pre-training heads.  Each event $x$ owned by user $u$ receives two scores: the sigmoided noise-detection logit $s_{\text{noise}}(x) = \sigma(g_\psi(x))$, and the negative cosine $s_{\text{proto}}(x) = -\cos(\phi(x), p_u)$ between the visit projection $\phi(x)$ and the entity prototype $p_u$ (high $=$ event does not match $u$'s pretraining-time prototype).
We convert each to a global rank in $[0, 1]$ across all test events and sum: $s(x) = \text{rank}(s_{\text{noise}}(x)) + \text{rank}(s_{\text{proto}}(x))$.
Rank fusion is robust to the different score scales of the two heads and avoids any tunable mixing weight.
No real \texttt{anomaly} labels are seen at any stage.

\subsection{ICU mortality prediction}
\label{appendix:eicu_baselines}
The four ICU-mortality baselines (STraTS, Raindrop, WaveGNN, SimMTM) are run from their released code on the corpora and patient-level $70/15/15$ splits of Appendix~\ref{appendix:split}. All four were retuned under the procedure of Appendix~\ref{appendix:hp_search}, since none ships defaults for corpora of this size.

\section{Variance quantification and cohort sensitivity}
\label{appendix:variance}

\subsection{Percentile bootstrap confidence intervals}
\label{appendix:bootstrap}
This appendix reports percentile bootstrap confidence intervals over test instances for the anomaly-detection results of Tables~\ref{tab:experiments:humob-anomaly} and~\ref{tab:experiments:auth}.

All intervals are $95\%$ percentile intervals over $10{,}000$ resamples ($1{,}000$ for NUMOSIM-LA). Event- and visit-level metrics resample events or visits; user- and agent-level metrics resample users or agents \emph{after} the scorer's own max-over-events aggregation.

Table~\ref{tab:bootstrap_ci} gives per-seed intervals for the LANL row of Table~\ref{tab:experiments:auth}, where the cohort is thinnest ($197$ positive events in $145{,}138$, and $38$ positive users in $1{,}559$) and the intervals are correspondingly wide.
Tuor-AAAI17's event-level AP of $.0112$ falls \emph{inside} the interval of three of the five seeds, so event-level AP is not separable, and \S\ref{sec:experiments:auth} claims only same-range parity. The weaker baselines are cleanly excluded --- DeepLog ($.0066$), NuLog ($.0059$) and LogGPT ($.0055$) lie outside all five intervals, LogBERT ($.0079$) and UniTraj$^{\dagger}$ ($.0075$) outside four of five.

\begin{table}[!htbp]
\centering
\scriptsize
\caption{Per-seed percentile bootstrap intervals for \ours{}'s event-level AP on LANL (the pre-train-only row of Table~\ref{tab:experiments:auth}), $10{,}000$ resamples over events. The final column is interval width relative to the point estimate.}
\label{tab:bootstrap_ci}
\begin{tabular}{cccc}
\toprule
Seed & Event AP & $95\%$ interval & width\,/\,point \\
\midrule
$0$ & $.0106$ & $[.0068,\ .0185]$ & $1.1\times$ \\
$1$ & $.0118$ & $[.0088,\ .0173]$ & $0.7\times$ \\
$2$ & $.0123$ & $[.0084,\ .0189]$ & $0.9\times$ \\
$3$ & $.0157$ & $[.0120,\ .0220]$ & $0.6\times$ \\
$4$ & $.0361$ & $[.0278,\ .0477]$ & $0.6\times$ \\
\bottomrule
\end{tabular}
\end{table}

Table~\ref{tab:bootstrap_seed_vs_instance} places the two variance sources side by side for the \ours{} cells of Table~\ref{tab:experiments:humob-anomaly}. The last column is $2\sigma$ across seeds divided by the mean instance-interval width over the five seeds: above $1$ means the training seed moves the number more than test-set resampling does, below $1$ the reverse.
On every cell the instance interval is the wider of the two, by $1.4$--$6.0\times$; the widest intervals are at agent level on Urban Anomalies, which has $1{,}000$ agents per city.

\begin{table}[!htbp]
\centering
\scriptsize
\setlength{\tabcolsep}{4pt}
\caption{Across-seed spread ($n{=}5$) against the mean instance-resampling interval width for the \ours{} cells of Table~\ref{tab:experiments:humob-anomaly}. Ratio is $2\sigma$ divided by the interval width.}
\label{tab:bootstrap_seed_vs_instance}
\begin{tabular}{lllccc}
\toprule
Dataset & Level & Metric & mean $\pm\ \sigma$ & interval width & ratio \\
\midrule
NUMOSIM-LA & visit & AP      & $.1950\pm.0098$ & $.0279$ & $0.70\times$ \\
NUMOSIM-LA & visit & Max F1  & $.2901\pm.0089$ & $.0352$ & $0.50\times$ \\
NUMOSIM-LA & agent & AP      & $.5764\pm.0155$ & $.0989$ & $0.31\times$ \\
NUMOSIM-LA & agent & Max F1  & $.6485\pm.0114$ & $.0866$ & $0.26\times$ \\
\midrule
UA-Berlin  & agent & AP      & $.4132\pm.0142$ & $.1710$ & $0.17\times$ \\
UA-Berlin  & agent & Max F1  & $.4261\pm.0237$ & $.1802$ & $0.26\times$ \\
\midrule
UA-Atlanta & agent & AP      & $.4141\pm.0231$ & $.1802$ & $0.26\times$ \\
UA-Atlanta & agent & Max F1  & $.4214\pm.0217$ & $.1696$ & $0.26\times$ \\
\bottomrule
\end{tabular}
\end{table}

\subsection{Wall-clock cost}
\label{appendix:wallclock}
All timings are from the runs behind the reported numbers, on one NVIDIA RTX 6000 Ada (48\,GB). Pre-training is once per dataset (once per seed on LANL) and is reused by every downstream task on that dataset; each seed fine-tunes both co-occurrence routes (\S\ref{sec:experiments}), and the table reports the selected one.

\begin{table}[!htbp]
\centering
\scriptsize
\caption{Wall-clock time, one RTX 6000 Ada. Pre-train times are each backbone's own accumulated runtime (LANL: mean over its five backbones); fine-tune times are the mean over the reported seeds of the selected route, for the most expensive task on that dataset. Pre-training early-stops, so ``epochs'' is what ran, not the budget.}
\label{tab:wallclock}
\begin{tabular}{lrrr}
\toprule
Dataset & Pre-train (h) & Epochs run & Fine-tune (h) \\
\midrule
NUMOSIM-LA           & $16.21$ & $36$  & $49.67$ \\
UA-Berlin            & $3.23$  & $586$ & $0.89$ \\
UA-Atlanta           & $2.45$  & $200$ & $2.53$ \\
Foursquare-Tokyo     & $19.01$ & $211$ & $3.99$ \\
Gowalla-Stockholm    & $8.61$  & $399$ & $2.65$ \\
Gowalla-Austin       & $4.52$  & $235$ & $1.11$ \\
LANL                 & $0.62$  & $121$ & --- \\
eICU-CRD             & $2.46$  & $29$  & $2.80$ \\
MIMIC-IV             & $2.78$  & $80$  & $1.06$ \\
\bottomrule
\end{tabular}
\end{table}

LANL has no fine-tune time because \ours{} is not fine-tuned there (Appendix~\ref{appendix:lanl_scoring}). NUMOSIM-LA has the longest end-to-end time, $16.21$\,h of pre-training and $49.67$\,h of fine-tuning; every other dataset completes end-to-end in under one day.

\subsection{Ablation protocol}
\label{appendix:ablation_protocol}
The protocol of Table~\ref{tab:experiments:ablation} differs from that of the per-task tables of \S\ref{sec:experiments} as follows, so the Full \ours{} row carries different numbers in the two.

\begin{enumerate}
\item Each column family uses one fixed co-occurrence route for all rows: active for mobility anomaly detection, social link, and the authentication log, and bypass for next-visit. The per-task tables select the route per seed on the validation criterion (\S\ref{sec:experiments}).
\item Cells with $\pm$ report the mean over five pre-training seeds with the fine-tuning seed fixed, except in row C6, which has no pre-training and varies the fine-tuning seed over five values, and for Full \ours{} on the authentication log, which is not fine-tuned (item 4). For Full \ours{} on mobility anomaly detection and next-visit, $\pm$ is $\sqrt{\sigma^2_{\text{pre}}+\sigma^2_{\text{fine}}}$, with $\sigma_{\text{fine}}$ taken over three and five fine-tuning seeds, respectively; elsewhere $\pm$ is the standard deviation over the varied seed. Cells without $\pm$ are single runs. The per-task tables vary the fine-tuning seed from one pre-trained checkpoint per dataset, except Table~\ref{tab:experiments:auth}, which uses five pre-trained backbones.
\item Social link is reported from the model score alone. Table~\ref{tab:experiments:social} late-fuses the model score with a logistic regression on handcrafted co-visitation features at a validation-selected mixing weight (Appendix~\ref{appendix:social_baselines}).
\item The authentication-log columns are scored at user level. Full \ours{} is scored without fine-tuning from its five pre-trained backbones, as in Table~\ref{tab:experiments:auth} (Appendix~\ref{appendix:lanl_scoring}); every other row is fine-tuned on LANL, C6 from scratch, so in these columns each arm also differs from Full in fine-tuning.
\item Rows C4 and C5 replace the denoising objective with next-token and masked-token prediction, respectively. For C4 on mobility anomaly detection and the authentication log, and for C5 on the authentication log, the backbone was pre-trained without $\mathcal{L}_{\text{prototype}}$ as well, and so differs from Full \ours{} in two components; the other C4 and C5 cells keep $\mathcal{L}_{\text{prototype}}$ and differ in the objective only. On the three arm--task pairs evaluated both ways, keeping $\mathcal{L}_{\text{prototype}}$ changes the variant's score by $-.003$ AP / $-.001$ H@$10$ (C4, social link), $-.003$ AP / $-.002$ H@$10$ (C5, social link), and $-.027$ AP / $-.024$ MaxF1 (C5, mobility anomaly detection).
\end{enumerate}

\subsection{LANL cohort-construction sensitivity}
\label{appendix:lanl_cohort}
The LANL cohort of Appendix~\ref{appendix:lanl} fixes a sampling seed and a per-user event cap. Table~\ref{tab:lanl_cohort_sensitivity} scores \ours{} without fine-tuning, as in Table~\ref{tab:experiments:auth}, on two perturbations of that construction: a different benign-user sample at the same cap, and the same sample at twice the cap. Rows $2$--$4$ each use one backbone pre-trained on that cohort with the same recipe; row $1$ repeats Table~\ref{tab:experiments:auth}.
Relative to row $2$, doubling the cap lowers all six metrics, and re-sampling the benign users lowers the three event-level metrics and user Max F1 while raising user AP and AUROC. Every value in rows $2$--$4$ lies within two standard deviations of the mean in Table~\ref{tab:experiments:auth}, except user Max F1 at twice the cap ($.2826$, $3.2$ standard deviations below).

\begin{table}[!htbp]
\centering
\scriptsize
\caption{LANL cohort-construction sensitivity, \ours{} scored without fine-tuning. Row $1$ is Table~\ref{tab:experiments:auth} (mean $\pm$ standard deviation over five backbones); rows $2$--$4$ use one backbone each.}
\label{tab:lanl_cohort_sensitivity}
\setlength{\tabcolsep}{4pt}
\begin{tabular}{lllllll}
\toprule
Sample seed / cap & \hc{Event AP} & \hc{Event AUROC} & \hc{Event Max F1} & \hc{User AP} & \hc{User AUROC} & \hc{User Max F1} \\
\midrule
$42$ / $500$ (Table~\ref{tab:experiments:auth}) & $.0173\std{.011}$ & $.8594\std{.052}$ & $.0683\std{.030}$ & $.2392\std{.039}$ & $.8611\std{.033}$ & $.4151\std{.042}$ \\
$42$ / $500$ & $.0264$ & $.9235$ & $.0858$ & $.2057$ & $.8668$ & $.3617$ \\
$202$ / $500$ & $.0089$ & $.8543$ & $.0367$ & $.2727$ & $.9146$ & $.3385$ \\
$42$ / $1000$ & $.0036$ & $.8233$ & $.0224$ & $.1730$ & $.8323$ & $.2826$ \\
\bottomrule
\end{tabular}
\end{table}

\subsection{Capacity-matched next-POI baselines}
\label{appendix:capacity_matched}
To separate the prototype mechanism from user-embedding capacity, we set the user-embedding dimensionality of LoTNext and MobTCast to $\dim(p_u){=}208$ on Gowalla-Austin, with three runs per method and a separately pre-trained \ours{} backbone. \ours{} reaches H@$10$ $.2579{\scriptscriptstyle\pm.0026}$ against LoTNext $.1951{\scriptscriptstyle\pm.0059}$ and MobTCast $.1595{\scriptscriptstyle\pm.0048}$. Replacing the InfoNCE of Eq.~\ref{eq:prototype} with a full $5{,}397$-way user-classification cross-entropy at the same capacity gives $.2502{\scriptscriptstyle\pm.0023}$.

\section{Extended related work}
\label{appendix:related_work_extended}

Each downstream task in \S\ref{sec:experiments} has a bespoke baseline lineage.  Table~\ref{tab:related_work_extended} compares each baseline against \ours\ along three axes: (i) \emph{input} --- what the model consumes per training example; (ii) \emph{supervision} --- the loss it trains under; (iii) \emph{cross-entity coupling} --- whether and how information from other entities enters the model.  \ours\ consumes the per-event tuple $(u, x, \tau, a, \delta)$ of \S\ref{sec:meses}, supervises through the dual denoising-plus-entity-prototype objective of \S\ref{sec:method}, and routes cross-entity signal through the co-occurrence attention axis (\S\ref{sec:method:arch}) and the shared prototype table (\S\ref{sec:method:user_loss}).

\begin{table}[!htbp]
\centering
\scriptsize
\setlength{\tabcolsep}{3pt}
\renewcommand{\arraystretch}{1.2}
\caption{Per-baseline comparison along three axes.  Family blocks group baselines benchmarked under each downstream task in \S\ref{sec:experiments}; the bottom row summarizes \ours\ for reference.}
\label{tab:related_work_extended}
\begin{tabular}{@{}p{0.13\linewidth} p{0.25\linewidth} p{0.38\linewidth} p{0.2\linewidth}@{}}
\toprule
Method & Input & Supervision & Cross-entity coupling \\
\midrule
\multicolumn{4}{@{}l}{\emph{Mobility anomaly detectors}} \\
ICAD~\citep{azarijoo2025icad}            & per-user visit sequence                       & SSL autoregressive next-event; anomaly from likelihood             & none (per-user) \\
CoBAD~\citep{wen2025cobad}               & per-user visits + collective context          & likelihood under a collective-behavior model                       & explicit, per-event likelihood \\
USTAD~\citep{wen2025uncertainty}         & per-user spatio-temporal visits               & supervised prediction with an uncertainty head                     & none (per-user) \\
DeepBayesic~\citep{duan2024back}        & per-user visits + population density          & statistical-likelihood + neural prediction                          & shared density prior \\
\midrule
\multicolumn{4}{@{}l}{\emph{Next-POI rankers}} \\
TAPT~\citep{xu2025tapt}                   & per-user visit history                        & joint next-POI + next-timestamp CE                                 & none \\
MobTCast~\citep{xue2021mobtcast}         & per-user visit history                        & next-POI CE + auxiliary trajectory forecasting                     & none \\
LoTNext~\citep{xu2024taming}             & per-user visit history                        & long-tail-aware re-weighted next-POI CE                            & none \\
GETNext~\citep{yang2022getnext}          & per-user history + global flow map            & next-POI CE over flow-map-augmented representation                 & pre-committed flow graph \\
COAST~\citep{xin2025coast}               & per-user check-ins + spatial POI graph        & next-POI CE + InfoNCE over augmented sequence pairs                & spatial kNN POI graph (GAT) \\
\midrule
\multicolumn{4}{@{}l}{\emph{Next-visit predictors (joint location + time)}} \\
TrajGPT~\citep{hsu2024trajgpt}           & per-user visit history                        & autoregressive POI CE $+$ GMM NLL on next-visit time delta          & none (per-user) \\
Mobility-LLM~\citep{gong2024mobilityllm} & per-user visit history wrapped as LLM prompt  & POI CE $+$ L1 on next-visit time delta (min), joint-$\alpha$ weighted & none (per-user) \\
THP~\citep{zuo2020thp}                   & per-user marked event stream                  & marked-Hawkes log-likelihood $+$ next-mark CE $+$ scaled time-MSE   & none (per-user) \\
A-NHP~\citep{yang2022anhp}               & per-user marked event stream                  & attentive neural Hawkes log-likelihood (single term)                & none (per-user) \\
\midrule
\multicolumn{4}{@{}l}{\emph{Social-link inferers}} \\
EBM~\citep{pham2013ebm}                  & co-visitation tuples                          & closed-form entropy-based pair score (no learned parameters)       & central observable \\
LBSN2Vec~\citep{yang2019revisiting}      & check-ins as $(u,t,x,a)$ hyperedges           & hypergraph-embedding loss with social-edge supervision             & hyperedge structure \\
H\textsuperscript{3}GNN~\citep{li2025heterogeneous} & user-POI heterogeneous hypergraph (hyperbolic) & friend-recommendation supervision over the hypergraph         & hypergraph topology \\
USRC~\citep{chu2025one}                  & per-user mobility + pair features             & cross-city social-relationship classification                      & pair-level coupling \\
\midrule
\multicolumn{4}{@{}l}{\emph{Enterprise-log anomaly detectors}} \\
DeepLog~\citep{du2017deeplog}            & per-host log-key sequence                     & LSTM next-token CE; anomaly from likelihood                        & none (single-stream) \\
Tuor-AAAI17~\citep{tuor2017deep}         & per-user multi-field log sequence             & summed CE across categorical fields                                & none (single-stream) \\
LogBERT~\citep{guo2021logbert}           & per-host log-key sequence                     & masked-token reconstruction ($+$ hypersphere)                      & none (single-stream) \\
NuLog~\citep{nedelkoski2020nulog}        & per-host log-token sequence                   & masked-token reconstruction with $[\text{CLS}]$-pooled anomaly head & none (single-stream) \\
LogGPT~\citep{han2023loggpt}             & per-host log-key sequence                     & GPT autoregressive next-token CE; anomaly from likelihood          & none (single-stream) \\
\midrule
\multicolumn{4}{@{}l}{\emph{ICU mortality predictors}} \\
STraTS~\citep{tipirneni2022strats}       & per-stay irregular event triples              & self-supervised forecasting pre-train $+$ supervised binary CE     & none (per-stay) \\
Raindrop~\citep{zhang2022graph}          & per-stay sensor multivariate time series      & dependency-graph message passing, supervised binary CE             & per-stay sensor graph \\
WaveGNN~\citep{hajisafi2025wavegnn}      & per-stay irregular multivariate time series   & wavelet-based GNN, supervised binary CE                            & per-stay sensor graph \\
SimMTM~\citep{dong2023simmtm}            & per-stay dense [$C$,$T$] time-series grid     & masked-reconstruction $+$ series-wise contrastive pre-train $+$ supervised binary CE & none (per-stay) \\
\midrule
\textbf{\ours{}}                         & per-event tuple $(u,x,\tau,a,\delta)$         & SSL pre-train (denoising + entity-prototype) $+$ task-specific fine-tune (self-supervised on anomaly tasks; supervised on next-POI / social / ICU mortality) & learned co-occurrence attention $+$ shared prototype \\
\bottomrule
\end{tabular}
\end{table}

CoBAD~\citep{wen2025cobad} is the closest baseline in spirit on the cross-entity axis, but its coupling enters through a per-event collective-behavior likelihood rather than a pre-training objective, so the cross-entity signal cannot be reused across downstream tasks.  GETNext~\citep{yang2022getnext} routes cross-entity signal through a global trajectory flow map computed once from the training set, whereas \ours's co-occurrence axis recomputes attention weights per visit from the current event window.  THP and A-NHP, the temporal-point-process baselines for next-visit prediction, model continuous-time event streams via attention-driven marked Hawkes intensities (with auxiliary mark-CE and time-MSE heads in THP) but operate per-user; \ours\ shares the same prediction target (next-event time and mark) yet learns its representations under a discriminative denoising objective rather than an event-likelihood loss, which the next-visit results show transfers to high H@$k$.  In the social-link family, EBM, LBSN2Vec, H\textsuperscript{3}GNN, and USRC all consume cross-entity structure as the central training signal; \ours\ instead learns its per-user prototypes during pre-training without any social-edge supervision, and the social-link head in \S\ref{sec:experiments:social} is a thin pair scorer over those prototypes rather than a graph network.  LogBERT~\citep{guo2021logbert} is the closest baseline to \ours's pre-train-plus-fine-tune paradigm; its single-stream restriction is the structural gap, since cross-entity coupling between users sharing hosts is an explicit signal on LANL that single-stream models cannot use.

\paragraph{Prototype-contrastive learning beyond mobility.}
Prototype-anchored contrastive losses appear elsewhere with a different supervisory commitment.
PCL~\citep{li2021pcl} alternates clustering of unsupervised image embeddings with a contrastive loss whose positive class is the cluster centroid; SWCC~\citep{gao2022swcc} augments InfoNCE on event sentences with a clustering branch whose centroids regularize semantically related events from being pulled apart by the contrastive negatives.
In both, the prototype is a \emph{latent} cluster centroid discovered from the data, and the assignment of an example to a prototype is itself an estimand.
\ours's $p_u$ is instead supervised by entity identity --- the assignment is given for free in every MESES dataset (\S\ref{sec:meses}, A2) --- and the same vector enters the input side of $f_\theta$, so contrastive shaping at the output and feature use at the input are coupled by construction.
On the place side of the (entity, context) pair, \citet{siampou2026mepois} learn POI-side prototypes from mobility data, the dual problem to ours.

\paragraph{Summary.}
Across all six families, \ours\ differs in two consistent ways: (i) supervision decomposes into a self-supervised pre-train (denoising plus entity-prototype) and a task-specific fine-tune --- self-supervised on the anomaly tasks (synthetic perturbations of $\eta$ at fine-tune; the LANL row skips the fine-tune and is scored from the pre-trained backbone, Appendix~\ref{appendix:lanl_scoring}) and supervised on the labeled tasks (next-POI, social link, ICU mortality) --- rather than a single end-to-end supervised stage; (ii) cross-entity coupling enters through learned attention and a shared prototype table rather than a pre-committed graph topology, a hand-built collective-behavior likelihood, or single-stream per-entity processing.

\section{Limitations and outlook}
\label{appendix:limitations}

\paragraph{Discriminative vs.\ likelihood-based supervision.}
\ours{} pre-trains representations for classification and ranking; on next-visit time prediction (Table~\ref{tab:experiments:next_visit}) it tops H@$10$/H@$20$ across all three LBSN city subsets but trails the LLM-based Mobility-LLM~\citep{gong2024mobilityllm} and the marked-Hawkes A-NHP~\citep{yang2022anhp} on T$\pm60$ --- the metric that rewards a calibrated density over event times.
For next-event time, hazard estimation, or any task whose primitive is a likelihood over the event distribution, TPP-style supervision remains the natural fit; a hybrid that adds a likelihood head on top of the \ours{} backbone is a clean direction for extending the framework to those tasks without losing the discriminative gains documented here.

\paragraph{Corruption operator is intentionally simple.}
The corruption operator $\eta$ resamples each flagged event's context $x_{u,t}$ uniformly from the bounding box of $\phi(\mathcal{X})$ (snapped to the nearest $x' \in \mathcal{X}$, with the activity $a_{u,t}$ replaced by the activity associated with $x'$) and resamples its timestamp $\tau_{u,t}$ uniformly within the window between adjacent events (\S\ref{sec:method:noise}); the activity $a_{u,t}$ thus shifts as a side-effect of the context-snap, and the duration $\delta_{u,t}$ shifts as a side-effect of the gap-preservation step on adjacent events.
This is a deliberately MESES-universal choice that requires only the (context, time) pair guaranteed by every MESES instance (\S\ref{sec:meses}); nonetheless it leaves A1's joint plausibility only partially constrained, since a context-conditional resampler --- e.g., time-of-day-aware context priors, near-but-atypical contexts as hard negatives, or marked corruption of the activity and duration channels --- would more tightly target the joint density over $(x, \tau, a, \delta)$ and is the most direct lever we are aware of for sharpening A1 alignment.
We did not run a corruption-design ablation along this axis, and view a systematic study of harder negatives --- including their interaction with $\mathcal{L}_{\text{prototype}}$ and the co-occurrence sub-layer --- as future work.

\paragraph{Co-occurrence route selected on validation, not learned.}\label{appendix:dispatch}
The co-occurrence route is chosen per seed on the validation criterion (\S\ref{sec:experiments}). It is active on every seed of NUMOSIM-LA anomaly detection and of social-link prediction, and bypassed on every seed of next-POI and next-visit prediction; it is bypassed on $4$ of $5$ seeds for UA-Atlanta and MIMIC-IV, $3$ of $5$ for eICU-CRD, and $2$ of $5$ for UA-Berlin. Learning the choice end-to-end, for example as a gate with a per-task or per-event coefficient, would remove the validation step.

\paragraph{Per-instance pre-training, not yet cross-instance.}
We pre-train one checkpoint per dataset --- six mobility datasets, the LANL authentication log, and the two EHR corpora --- which validates that the same recipe transfers across MESES instances but does not yet exploit the obvious next gain: a single backbone trained on a unified corpus of MESES instances.
Characterizing when a shared substrate is rich enough for cross-instance transfer (mobility-only, mobility$+$auth, mobility$+$EHR), and identifying whether the prototype table or the substrate embedding is the cross-instance bottleneck, are the directions we view as most likely to extend the framework beyond the per-instance checkpoints studied here.
A related deployment-time limitation is cold start on entities unseen during pre-training: the prototype table is keyed by training-set users, so a freshly observed user must either inherit a generic initialization or accumulate enough events to be fine-tuned in, and we have not characterized the few-shot regime where this matters most.

\paragraph{Single-run cells in the ablation table.}
In Table~\ref{tab:experiments:ablation}, $30$ of the $64$ cells are means over five seeds (Appendix~\ref{appendix:ablation_protocol}); the other $34$ are single runs, carry no error bar, and are not tested for significance.

\paragraph{Scope of the clinical evaluation.}
Table~\ref{tab:experiments:eicu} reports eICU-CRD ($197{,}307$ stays) and MIMIC-IV ($94{,}407$ stays); both are retrospective US intensive-care databases, and both are scored on the same binary outcome, in-hospital mortality. We therefore do not test whether the recipe transfers to non-ICU care, to non-US health systems, to longitudinal outcomes, or across sites within a corpus: our splits are patient-level but not site-held-out, so a hospital seen in training may also appear at test.

\paragraph{Substrate embedding $\phi$ is fixed, not learned.}
For the two non-mobility instances we obtain $\phi$ by running $2$D UMAP on the user--host (LANL) and context--stay (eICU) co-occurrence matrix at fixed hyperparameters ($n_{\text{neighbors}}{=}30$, cosine distance; Appendices~\ref{appendix:lanl}, \ref{appendix:eicu}) and freezing the result before pre-training.
The framework treats $\phi$ as part of the schema (\S\ref{sec:meses}), so any unsupervised geometry over $\mathcal{X}$ that respects co-occurrence would satisfy the formal requirements; we did not sweep $n_{\text{neighbors}}$, distance metric, or the embedding dimensionality, and we did not test an end-to-end learned $\phi$ that is updated jointly with the encoder under $\mathcal{L}$ (Eq.~\ref{eq:loss}).
A learned $\phi$ would close the loop between the substrate geometry that $\eta$'s context perturbation samples from (\S\ref{sec:method:noise}) and the representations the encoder builds on top of it, and is the most direct extension for the non-mobility instances.

\paragraph{Evidence for the identity-shortcut constraint.}
The shared-prototype design relies on $\mathcal{L}_{\text{noise}}$ to prevent an identity shortcut from the input-side $p_u$ to its supervisory copy (Appendix~\ref{appendix:identity_shortcut}); ablation row C2 (Table~\ref{tab:experiments:ablation}) shows the constraint is empirically binding, but the evidence is behavioral rather than structural.
We did probe the representations directly --- linear probes recovering the entity identity from each layer's $\mathcal{H}_{u,t}$, with frozen prototypes and layer by layer, on NUMOSIM-LA and UA-Berlin --- and found \emph{no} evidence that the prototype embeddings encode behavioral structure beyond identity.
What remains untried is an \emph{architectural} barrier: masking the residual on the entity-token path for selected layers, or a stop-gradient on the $p_u$ pathway past a chosen depth, would block the trivial copy operationally rather than by training pressure, and an ablation that toggles that barrier would distinguish ``the loss happens to suppress the copy'' from ``the architecture forbids it.''

\section{Broader impacts}
\label{appendix:broader_impacts}

\paragraph{Positive impacts.}
\ours{} consolidates anomaly detection, next-visit prediction, social-link inference, and clinical mortality prediction under a single self-supervised pre-training framework, reducing the engineering cost of building task-specific models for sparse multi-entity event streams. On the mobility side, better next-visit and next-POI prediction supports downstream applications such as demand forecasting for transit planning, congestion aware routing, and location-based recommendations that surface relevant places to users without requiring them to specify their intent in advance. 
Label-free anomaly detection also recovers lateral-movement-shaped activity in enterprise authentication logs (\S\ref{sec:experiments:auth}), a useful capability for organizations that can deploy it on their own logs without requiring labelled red-team events.
On the clinical side, the AP and Sens@Sp.9 gains on the rare-positive eICU mortality task (\S\ref{sec:experiments:eicu}) land at the operating point most relevant to early-warning use, where the cost of a missed positive dominates.

\paragraph{Negative impacts.}
\ours{} improves pre-trained representations for sparse multi-entity event streams, and like any improvement in general-purpose representations, its gains are not intrinsically tied to any specific use. The Axiom A2 and A3 mechanisms that make the framework accurate (entity-distinguishing information, co-occurrence structure) are also the mechanisms that any downstream task exploiting entity- or population-level structure would benefit from, including uses beyond those we evaluated. As with other pre-training advances, practitioners deploying \ours{} on new data should apply the oversight standards appropriate to that setting.

\paragraph{Mitigations.}
All experiments use datasets publicly released by their original owners under their own anonymization and licensing decisions (Appendix~\ref{appendix:datasets}); we introduce no new identifiers and redistribute no raw data. Practitioners deploying \ours{} on new data should validate its behavior on that data before use and apply the access controls that govern the underlying records to the model's per-entity outputs.

\end{document}